\documentclass[11pt]{article}

\usepackage[utf8]{inputenc}
\usepackage[T1]{fontenc}
\usepackage[margin=1in]{geometry}
\usepackage{hyperref}
\usepackage{url}
\usepackage{booktabs}
\usepackage{amsfonts}
\usepackage{nicefrac}
\usepackage{microtype}
\usepackage{graphicx}
\usepackage[numbers]{natbib}
\usepackage{subfigure}
\usepackage{multirow}
\usepackage{amsmath,bm}
\usepackage{algorithm}
\usepackage{algorithmic}
\usepackage{enumitem}
\usepackage{bbm}
\usepackage{float}
\usepackage{pifont}
\usepackage{caption}
\usepackage{xcolor}

\newcommand{\rom}[1]{\uppercase\expandafter{\romannumeral #1\relax}}
\newcommand{\cmark}{\ding{51}}%
\newcommand{\xmark}{\ding{55}}%

\title{$P^2$GNN: Two Prototype Sets to boost GNN Performance}

\author{
  Arihant Jain \quad Gundeep Arora \quad Anoop Saladi \quad Chaosheng Dong \\
  Amazon, India \\
  \texttt{\{arihanta, gundeepa, saladias, chaosd\}@amazon.com}
}

\date{}

\begin{document}
\maketitle

\begin{abstract}
Message Passing Graph Neural Networks (MP-GNNs) have garnered attention for addressing various industry challenges, such as user recommendation and fraud detection. However, they face two major hurdles: (1) heavy reliance on local context, often lacking information about the global context or graph-level features, and (2) assumption of strong homophily among connected nodes, struggling with noisy local neighborhoods. To tackle these, we introduce $P^2$GNN, a plug-and-play technique leveraging prototypes to optimize message passing, enhancing the performance of the base GNN model. Our approach views the prototypes in two ways: (1) as universally accessible neighbors for all nodes, enriching global context, and (2) aligning messages to clustered prototypes, offering a denoising effect. We demonstrate the extensibility of our proposed method to all message-passing GNNs and conduct extensive experiments across 18 datasets, including proprietary e-commerce datasets and open-source datasets, on node recommendation and node classification tasks. Results show that $P^2$GNN outperforms  production models in e-commerce and achieves the top average rank on open-source datasets, establishing it as a leading approach. Qualitative analysis supports the value of global context and noise mitigation in the local neighborhood in enhancing performance.
\end{abstract}


\section{Introduction}
\label{sec:intro}

Graph structured data plays a crucial role for depicting intricate relationships and interactions across various domains, such as social networks, chemistry, and biology. Nodes within a graph typically harbor rich features that offer nuanced insights. Graph Neural Networks (GNNs) have garnered significant traction due to their adeptness in tasks like node classification and recommendation.  Notable techniques like ChebNet~\cite{cnn_graph}, GCN~\cite{kipf2016classification}, GAT~\cite{velivckovic2017attention}, and GraphSAGE~\cite{hamilton2017inductive} rely on message passing among neighboring nodes, albeit with distinct approaches to feature transformation and message aggregation.
GNNs have found practical applications in fraud detection~\cite{fraud1, fraud2}, identifying complementary/supplementary products~\cite{jian2022multi,yan2022personalized,virinchi2023recommending}, etc., underscoring their potential to augment user experience and drive tangible business outcomes in industry.

However, existing GNN techniques face two significant limitations. First, GNNs heavily rely on local context, \textbf{lacking information about global context} or graph-level features. Second, message passing GNNs assume strong homophily among connected nodes, implying that linked nodes share similar characteristics or labels \cite{zhu2020beyond,luan2020complete,liu2021non,chien2021adaptive,zhu2020graph,yan2021two,ma2021homophily,li2022finding}. Consequently, they encounter difficulties when dealing with heterophilous nodes, where neighborhood information fails to reinforce positive signals in the presence of \textbf{noisy local neighborhoods}. These challenges often result in suboptimal performance of GNNs on certain datasets, sometimes even underperforming simple Multi-Layer Perceptron (MLP) baselines. For example, Graph Attention Network (GAT) models exhibit a 2.6\% performance decline compared to the MLP baseline on the Film dataset \cite{pei2020geom} (36.0\% vs. 38.6\% \cite{li2022finding}), attributable to the dataset's noisy signal and lack of global context.
Such challenges are prevalent in industry applications like fraud detection \cite{fraud1,fraud2}, where malicious users are connected to multiple benign users, thus rendering their neighborhood seemingly benign and impeding detection efforts. Addressing these limitations is paramount for enhancing the performance and practical utility of GNNs.

To mitigate the lack of global context information, we propose a straightforward yet effective method involving the creation of pseudo-nodes, or prototypes, which \textbf{encapsulate global-level information}. These prototypes establish dense connections with all nodes in the graph, enabling nodes to access information regarding features of nodes with similar characteristics or labels, thereby fostering awareness of global context. Compared to approaches involving the augmentation of graph density by adding edges between nodes, our method offers enhanced scalability ($\mathcal{O}(K*|V|)$ vs. $\mathcal{O}(|V|^2)$ for $K$ prototypes and $|V|$ with $K \ll |V|$), rendering it particularly well-suited for industry applications.

To address the challenge of noisy local neighborhoods, we draw inspiration from prototypical networks \cite{snell2017prototypical}. Prototypical networks offer a noise-reducing mechanism via hard-clustering, wherein cluster centers or prototypes eliminate intra-cluster variance. Our proposal involves aligning aggregated messages to such prototypes, aiming to enhance message passing by mitigating input noise to the aggregation function. This \textbf{denoising effect} is able to diminish the influence of irrelevant neighbor messages on downstream tasks.
\begin{figure}%{r}{0.5\linewidth}
    \centering
    \includegraphics[width=0.5\textwidth]{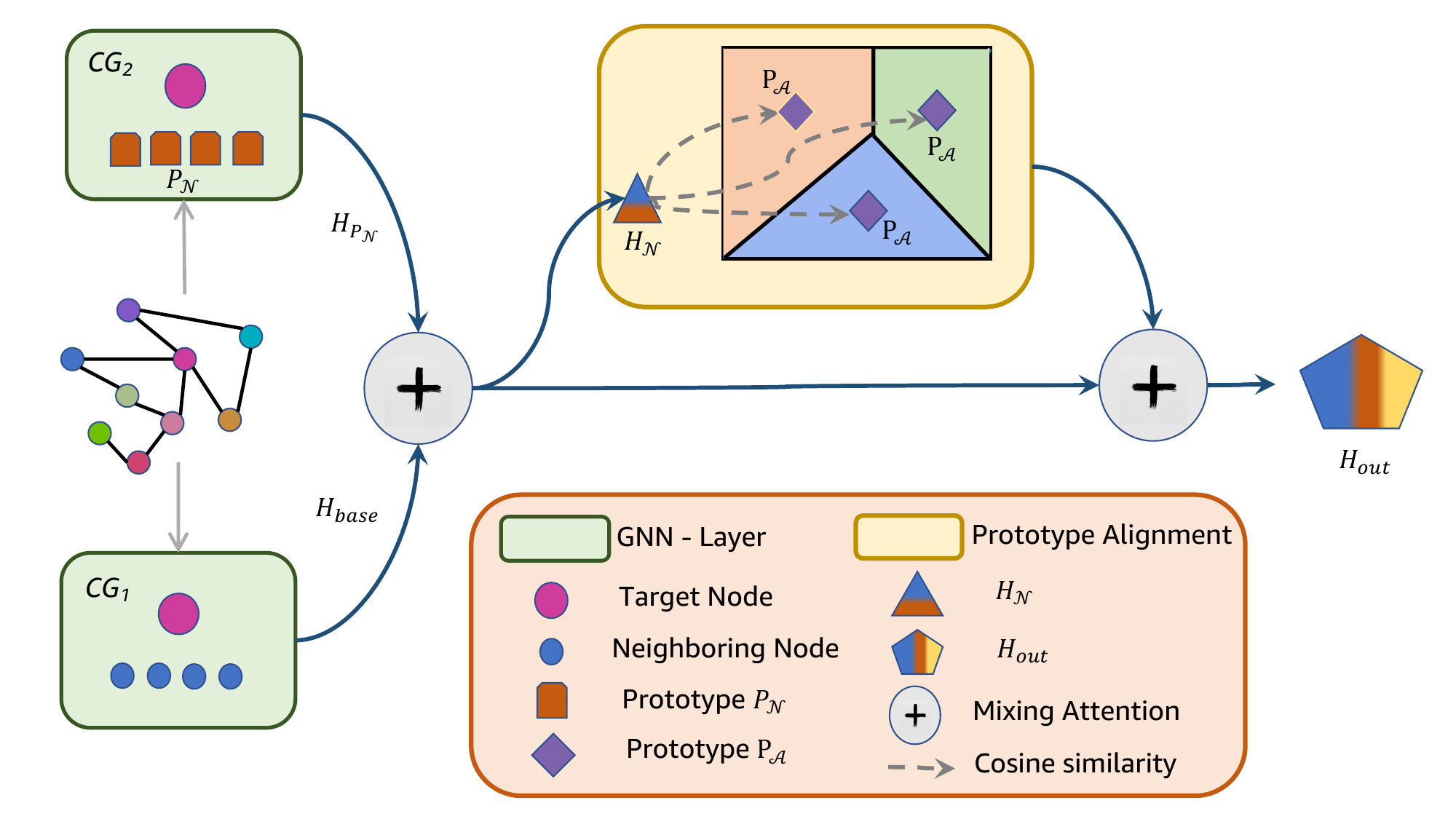}
    \caption{Schematic of \textbf{$P^2$GNN}. Two computational graphs for each target node: one with neighbouring nodes from source graph ($CG_1$) and another with learnable $K_\mathcal{N}$ class-independent prototypes $P_\mathcal{N}$ as neighbouring nodes ($CG_2$). GNN embeddings of target node obtained via message passing from both   ($H_{base}$ from $CG_1$, $H_{P_\mathcal{N}}$ from $CG_2$ as in Eq.~\eqref{eqn:fwd}) are mixed to form $H_\mathcal{N}$ (Eq.~\eqref{eqn:hnl_attn}). $H_\mathcal{N}$  is then aligned with $K_\mathcal{A}$ learnable class-independent $P_\mathcal{A}$ prototypes to create $H_\mathcal{A}$ (Eq.~\eqref{eqn:align_sf}), followed by mixing with $H_\mathcal{N}$ for final output $H_{out}$ (Eq.~\eqref{eqn:align_attn}).}
    \label{fig:schematic}
\end{figure}

In summary, we identified two major limitations in existing GNN frameworks and addressed them by utilizing prototypes. While using prototypes in GNNs may seem trivial due to extensive research on prototype-based methods, to the best of our knowledge, we are the first to employ them to tackle these GNN specific challenges.
Our novel contributions, summarized in Fig.~\ref{fig:schematic}, are three-fold:

\begin{itemize}[noitemsep,topsep=0pt,leftmargin=*]
    \item \textbf{Prototypes as Neighbors} [$P_{\mathcal{N}}$]: We introduce a novel approach that leverages prototypes as pseudo-nodes, providing their messages to all nodes in the graph, thereby offering global context.
    \item \textbf{Prototypes for Message Alignment} [$P_{\mathcal{A}}$]: We propose aligning messages to clustered prototypes via attention, thereby reducing noise during message passing, yielding a denoising effect.
    \item \textbf{Empirical Performance Boost}: Our methodology enhances GNN performance across 16 public benchmark datasets and 2 proprietary e-commerce datasets, showcasing improvements in two downstream tasks. Through an in-depth qualitative analysis of embeddings, we evaluate the incremental efficacy of each proposed component, firmly endorsing the utilization of prototypes.
\end{itemize}

\section{Related Work}
\label{sec:related}
Existing GNN architectures, such as GCN~\cite{kipf2016classification}, GAT~\cite{velivckovic2017attention}, and GraphSage~\cite{hamilton2017inductive}, have made significant strides in graph machine learning tasks. However, their reliance on message passing from neighbors can adversely affect the performance of noisy heterophilous nodes (nodes connected to nodes from different classes). Approaches to address this challenge can be classified into three main directions: (1) enhancing input features~\cite{hpgmn,chen2022memory,hou2019measuring,abu2019mixhop,pei2020geom}, (2) incorporating heterophily-aware aggregation methodologies~\cite{zhu2020beyond,elmoataz2008nonlocal,chien2021adaptive,bo2021beyond,du2022gbk,udgnn,he2021bernnet,suresh2021breaking}, and (3) improving message aggregation from neighbors~\cite{yang2021diverse,luan2022revisiting,lim2021large,klicpera2018predict}. However, none of these techniques harnesses prototypes to enhance GNN performance, with their primary focus predominantly on the node classification task. Additionally, we focused specifically on message passing GNNs in this work. Therefore, we chose to exclude spectral and geometric GNNs from comparison as they are not directly applicable to our case.

\textbf{Prototypes-based GNNs}: Recent research has centered on harnessing prototypes within GNN layers for various tasks on graph data. A considerable focus has been on enhancing GNN explanations \cite{ragno2022prototype,shin2022page,dai2022towards,zhang2022protgnn}. A considerable focus has been on enhancing GNN explanations \cite{ragno2022prototype,shin2022page,dai2022towards,zhang2022protgnn}, with prototypes solely used for improving the explainability of GNN outcomes and not for improving downstream task performance. Another work \cite{Ren_2023} focuses on the GNN recommendation task based on user-item interactions and employs contrastive learning method using latent representations.

However, some works have explored alternative uses for prototypes as well. DPGNN \cite{wang2021distance} introduced supervised prototypes but solely for imbalanced node classification tasks. HP-GMN \cite{hpgmn} proposed integrating graph-level memory units (akin to prototypes) at the final classification layer to augment performance. This approach, however, is not directly applicable to GNN because the memory units do not directly influence the message passing step. 

Closely related to our work is ProtoGNN \cite{dongprotognn}, which utilizes prototypes for pseudo-node message passing to enhance performance. However, ProtoGNN's approach differs significantly from ours. Their prototypes are derived from an independently trained MLP model that necessitates a subset of crucial training data, whereas our simultaneously learned prototypes require no specific data during the pre-training stage. Moreover, ProtoGNN's experiments are confined to binary classification tasks due to the linear increase in the number of required prototypes with the number of classes, rendering it impractical for industry applications. In contrast, $P^2$GNN remains agnostic to class counts and tasks, emphasizing diversity in the feature space for prototypes. This approach affords the model flexibility in determining the number of prototypes per class without imposing an equal-representation constraint. These pivotal distinctions are summarized in Table~\ref{table:related}.

\begin{table}[H]
\caption{$P^2$GNN vs. Other Prototype-based Methods}
\centering
\resizebox{0.48\textwidth}{!}{
\begin{tabular}{c|cccc}
    \hline
    {\bf Criterion} & {\bf DPGNN} & {\bf HP-GMN} & {\bf ProtoGNN} & {\bf $P^2$GNN} \\ \hline
    GNN-based & \cmark &\xmark & \cmark & \cmark \\ 
    No Pre-training & \cmark & \xmark&\xmark & \cmark \\ 
    Class independent &\xmark & \cmark &\xmark & \cmark \\ 
    Task-agnostic & \xmark & \xmark& \xmark& \cmark \\ \hline
\end{tabular}
}
\label{table:related}
\end{table}

\section{Methodology}

This section describes the two prototypes constituting the $P^2$GNN framework. Both prototypes are optimized for the downstream task using the same set of loss functions. Each component is designed for independent utilization and can be extended to all message-passing GNNs. The integration methodology of $P^2$GNN with a generic GNN layer is depicted in Fig.~\ref{fig:schematic} and Algorithm~\ref{alg:additionP2}, elucidating the introduced prototypes and their leveraging process. Table~\ref{tab:notation} enumerates the notations employed throughout the paper.

\begin{table}[!htb]
  \caption{List of Notations}
  \label{tab:notation}
  \centering
  \begin{tabular}{ll}
    \toprule
    \textbf{Notation}     & \textbf{Description}  \\
    \midrule
    $\mathcal{G} = (\mathcal{V}, \mathcal{E})$ & undirected graph without self-loops \\
    $\mathcal{V} = \{v_i\}_{i=1}^V$, $V$ & set of nodes, number of nodes \\
    $\mathcal{E} \subseteq \mathcal{V} \times \mathcal{V}$ & set of edges \\
    $A \subseteq \{0,1\}^{V \times V}$ & adjacency matrix\\
    $X \in \mathbb{R}^{V \times d_0}$ & initial node feature matrix\\
    $K_{\mathcal{N}}$ & Number of learnable $P_{\mathcal{N}}$ prototypes  \\
    $K_{\mathcal{A}}$ & Number of learnable $P_{\mathcal{A}}$ prototypes  \\
    $\mathcal{N}_i$ & neighborhood of node $v_i$\\
    $H^{l}$ & node representation matrix in the $l$-th layer\\
    $h_i^{l}$ & $i$-th row in the embedding vector of node $v_i$  \\ 
    $D$; $D_{ii}= \sum_{j=1}^V{A_{ij}}$ & diagonal matrix\\
    $\tilde{A} = A + I_n$ & adjacency matrix with self loops\\
    $\tilde{D} = D + I_n$ & diagonal matrix with self loops\\
    $\hat{{A}} = \tilde{D}^{-\frac{1}{2}}\tilde{A}\tilde{D}^{-\frac{1}{2}}$ &  normalized adjacency matrix\\
    $h$ & hidden dimension of the GNN layer\\
    $P_{\mathcal{N}} \in \mathbb{R}^{K_\mathcal{N} \times d_0}$ & Prototypes as Neighbours\\
    $P_{\mathcal{A}} \in \mathbb{R}^{K_\mathcal{A} \times h}$ & Prototypes for Message-Alignment\\
    \bottomrule
  \end{tabular}
\end{table}

\begin{small}
\begin{algorithm}[htbp]
\caption{Addition of $P^2$GNN to generic GNN model}
 \label{alg:additionP2}
 \begin{algorithmic}[1]
  \STATE  {\bf Input:} Graph $\mathcal{G}=(\mathcal{V},\mathcal{E})$ with adjacency $A$ and features $X$, number of prototypes $K_{\mathcal{N}}$, $K_{\mathcal{A}}$, $GNN^l$ for $l^{th}$ layer
  \STATE \texttt{\# Initialization}
  \STATE Initialise for each layer $l \in L$, GNN layer parameters ($GNN^l_{base}$, $GNN^l_{P}$) , $K_{\mathcal{N}}$ Prototypes $P_{\mathcal{N}}$ and  $K_{\mathcal{A}}$ Prototypes $P_{\mathcal{A}}^l$
  \STATE \texttt{\# Expansion of adjacency matrix for $P_{\mathcal{N}}$}
  \STATE $A_{base} = \left(\begin{array}{@{}c|c@{}}
    A & 0 \\\hline
    0 & 0 
\end{array}\right)$, $A_P = \left(\begin{array}{@{}c|c@{}}
    0 & 1 \\\hline
    0 & 0 
\end{array}\right)$ $\in \{0,1\}^{(V+K_{\mathcal{N}}) \times (V+K_{\mathcal{N}})}$
%   \STATE $A_P = \left(\begin{array}{@{}c|c@{}}
%     0 & 1 \\\hline
%     0 & 0 
% \end{array}\right)$ $\in \{0,1\}^{(|\mathcal{V}|+K_{\mathcal{N}}) \times (|\mathcal{V}|+K_{\mathcal{N}})}$
  
  \STATE Get normalization adjacency matrix $\hat{A}_{base}$ and $\hat{A}_{P}$
  \STATE $H^0 = 
  \left(\begin{array}{@{}c@{}}
    X \\\hline
    P_{\mathcal{N}}
\end{array}\right)$ $\in \mathbb{R}^{(V+K_{\mathcal{N}})\times d_0}$
  \FOR{$l=1$ to $L$}
  \STATE \texttt{\# Messages from both neighbours and Prototypes($P_{\mathcal{N}}$) }
  \STATE $H_{base}^l = GNN^l_{base} (\hat{A}_{base}, H^{l-1})$ using Eq.~\eqref{eqn:fwd}
  \STATE $H_{P_{\mathcal{N}}}^l = GNN^l_{P} (\hat{A}_{P}, H^{l-1})$ using Eq.~\eqref{eqn:fwd}
  \STATE $H_{\mathcal{N}}^l = mixing(H_{base}^l, H_{P_{\mathcal{N}}}^l$) using Eq.~\eqref{eqn:hnl_attn}%\texttt{\chaosheng{say something like see eq.2} }
  \STATE \texttt{\# Message-Alignment using $P_{\mathcal{A}}^l$}
  \STATE Obtain $H_{\mathcal{A}}^l$ using Eq. \eqref{eqn:align_sf} 
  \STATE Obtain $H_{out}^l$ using Eq. \eqref{eqn:align_attn}
  \STATE $H^l = H_{out}^l$
  \ENDFOR
\STATE {\bf return $H^L$} 
\end{algorithmic}

\end{algorithm}

\end{small}

\subsection{Prototypes as Neighbours}
\label{sec:pan}

We propose the introduction of $K_\mathcal{N}$ prototypes ($P_\mathcal{N}$) as neighbors to each existing node in the graph, aiming to capture global-level information. These prototypes function akin to additional nodes with directed edges to every other node in the graph, except for connections among themselves. They actively engage in the message passing process by dispatching messages to existing nodes but do not receive any. The representations of these prototypes can be conceptualized as pseudo-node features, distinct for each layer.

We formulate the representation of prototype $P_\mathcal{N}$ as $K_\mathcal{N} \times d_0$ trainable parameters, where $d_0$ denotes the dimension of input features $X$. We concatenate the node and prototype representations to form $H^0 = \begin{pmatrix}
    X \\ \hline
    P_{\mathcal{N}}
\end{pmatrix} \in \mathbb{R}^{(V+K_{\mathcal{N}}) \times d_0}$. To perform normal message passing, we extend the adjacency matrix $A$ to $A_{base} = \left(\begin{array}{@{}c|c@{}}
    A & 0 \\\hline
    0 & 0 
\end{array}\right) \in \{0,1\}^{(V+K_{\mathcal{N}}) \times (V+K_{\mathcal{N}})}$ using a GNN architecture. For prototype message passing, we utilize the same GNN architecture but with different weight parameters. Hence, we define the adjacency matrix for passing messages from prototypes to nodes as $A_P = \left(\begin{array}{@{}c|c@{}}
    0 & 1 \\\hline
    0 & 0 
\end{array}\right)$ $\in \{0,1\}^{(V+K_{\mathcal{N}}) \times (V+K_{\mathcal{N}})}$. We compute GNN embeddings with weight matrices $W^{l}_{base}$ and $W^{l}_{P_{\mathcal{N}}}$ using the normalized adjacency matrices $\hat{A}_{base}$ and $\hat{A}_P$ as below.
\begin{equation}
\label{eqn:fwd}
\begin{gathered}
    H_{base}^{l} = \sigma(\hat{A}_{base} H^{l-1}W^{l}_{base}), ~H_{P_\mathcal{N}}^{l} = \sigma(\hat{A}_{P} H^{l-1}W^{l}_{P_{\mathcal{N}}}) 
\end{gathered}
\end{equation}

Next, we learn the contribution of each representation using mixed attention of messages from both sources to generate the output embedding $H^{l}$ for layer $l$.  Post transformation~\eqref{eqn:fwd}, the aggregation via mixed attention of messages from different sources result in $H_{\mathcal{N}}^{l}$. 
\begin{equation}
\label{eqn:hnl_attn}
\begin{gathered}
         \tilde{\alpha}_{base}^l = \texttt{Sigmoid} \left({H}^{l}_{base} \tilde{W}^{l}_{base}\right),
         ~\tilde{\alpha}_{P_\mathcal{N}}^l = \texttt{Sigmoid} \left({H}^{l}_{P_\mathcal{N}} \tilde{W}^{l}_{P_\mathcal{N}}\right)\\
     \left[{\alpha}_{base}^l, {\alpha}_{P_\mathcal{N}}^l \right] 
    = \texttt{Softmax}\left((\left[\tilde{\alpha}_{base}^l,\tilde{\alpha}_{P_\mathcal{N}}^l\right]/T) W_\text{Mix}^l \right) \\
    {H_{\mathcal{N}}^{l}} = \sigma\left( \texttt{DIAG}(\alpha_{P_\mathcal{N}}^l){H}^{l}_{P_\mathcal{N}} + \texttt{DIAG}(\alpha_{base}^l){H}^{l}_{base}\right)
\end{gathered}
\end{equation}

The weights $\tilde{W}^{l}_{base}$, $\tilde{W}^{l}_{P_\mathcal{N}}$, $W_{\text{Mix}}^l$, and temperature $T$ are the new parameters introduced by our prototype addition. Following \cite{luan2022revisiting}, we set $T$ to number of channels ($T$ = 2).

The prototypes capture a global-level view of node features, and their induced connectivity to all nodes provides a mechanism to combine this information with local neighborhood information. The mixed attention mechanism determines the proportion of information needed from both sources to optimize for downstream task. This methodology is extensible to other GNN architectures, as long as the node representation and prototype-node message passing are carefully integrated. We summarize this mechanism in Fig.~\ref{fig:schematic} (left) and lines 9-12 
 in Algorithm~\ref{alg:additionP2}.

\subsection{Prototypes for Message-Alignment}
\label{sec:pma}
To mitigate the issue of noisy local neighborhoods, we draw upon the ideas of prototypical networks \cite{snell2017prototypical}. They present a noise-eliminating mechanism through hard-clustering, as the cluster centers or prototypes eliminate intra-cluster variance. 
Drawing motivation from prototypical networks \cite{snell2017prototypical}, we propose that alignment to such prototypes can aid in obtaining well-separated representation learning therefore improve message passing by reducing input noise. This denoising effect can potentially minimize the impact of irrelevant neighbor messages for the downstream task.

We curate another set of $K_\mathcal{A}$ learnable prototypes, $P_\mathcal{A}$. For a message $H_{\mathcal{N}}^l$ to be passed from $u$ to $v$, the alignment operation finds the most well-aligned prototypes among $P_\mathcal{A}$. This effectively clusters the input messages, reducing variation compared to vanilla representations. 
This is achieved via a simple attention mechanism to obtain the representation $H^l_{\mathcal{A}}$ as follows.
\begin{eqnarray}
\label{eqn:align_sf}
    H^l_{\mathcal{A}} = \texttt{Softmax}(P_{\mathcal{A}}^l {H_{\mathcal{N}}^l}^T) P_{\mathcal{A}}^l 
\end{eqnarray}
Eq.~\eqref{eqn:align_sf} can be viewed as a special case of the Query-Key-Value attention method used in transformers \cite{vaswani2017attention}, where query = $P^l_A$, key = $H^l_N$, and value = $P^l_A$. The softmax operation, $\texttt{Softmax}(P^l_A{H^l_N}^T)$, calculates the contribution of each prototype to the final aligned message, which is then multiplied with its respective learnt prototype embedding to create the final aligned message $H_A^l$. 

Consequently, the convex combination of the original message $H_{\mathcal{N}}^l$ and the aligned message $H^l_{\mathcal{A}}$ is then used to obtain the final representation ${H_{out}^{l}}$. 
\begin{equation}
\begin{gathered}
\label{eqn:align_attn}
        \tilde{\alpha}^l = \texttt{Sigmoid} \left({H_\mathcal{N}^{l}} \tilde{W}^{l}\right), ~\tilde{\alpha}_{\mathcal{A}}^l = \texttt{Sigmoid} \left({H}^{l}_{\mathcal{A}} \tilde{W}^{l}_{\mathcal{A}}\right)\\
\left[{\alpha}^l, {\alpha}_{\mathcal{A}}^l \right] = \texttt{Softmax}\left((\left[\tilde{\alpha}^l,\tilde{\alpha}_{\mathcal{A}}^l\right]/T) W_\text{Mix}^l \right)\\
{H_{out}^{l}}  = \sigma\left(\texttt{DIAG}(\alpha^l){H_{\mathcal{N}}^{l}} + \texttt{DIAG}(\alpha_{\mathcal{A}}^l){H}^{l}_{\mathcal{A}} \right)
\end{gathered}
\end{equation}
Similar to previous mixing attention, the weights $\tilde{W}^{l}$, $\tilde{W}^{l}_{\mathcal{A}}$, $W_{\text{Mix}}^l$, and temperature $T$ (set to 2) are the new parameters introduced by our prototype addition. This process is summarized in Fig.~\ref{fig:schematic} (right) and lines 13-16 in Algorithm~\ref{alg:additionP2}.

\subsection{Prototypes Representation Optimization}
\label{sec:pro}
Given the two different prototype integrations ($P_\mathcal{N}$ and $P_\mathcal{A}$), we outline the desirable properties for the prototype representations to optimize task performance. These properties apply to both $P_\mathcal{N}$ and $P_\mathcal{A}$, albeit with varying degrees based on the dataset. Each property is enforced by a corresponding loss function, and the weighted combination of these losses, along with the downstream task loss, constitutes the final loss function. We introduce below three major loss functions: Alignment, Diversity, and Sparsity.

\textbf{Alignment}: Prototypes should have a hard-clustering effect and be well aligned with  features that will attend to them, i.e., prototypes should be representative of the input. To achieve this, we select the prototype that has the maximum cosine similarity to input embedding vector $h$. By aggregating maximum similarities over $h$, we can optimize prototypes to maximize their representativeness. 
\begin{equation}
\label{eqn:alignment_loss}
    \mathcal{L}_{alignment} (P) = - \sum_{h}\max_{p \in P}{cos(h, p)}
\end{equation}
This aligns with the aim that the attention of the representation with the prototype set should be maximally focused on the best-aligning prototype, reinforcing the hard-clustering objective. This is similar to the objective  optimized by the $K$-Means~\cite{arthur2006k,banerjee2005clustering}. However, applying $K$-Means to learn the prototypes is not practical as it would significantly increase the training time.

\textbf{Diversity}: With Eq.~\eqref{eqn:alignment_loss}, it is possible for the prototype representation to collapse to the mean of input features ($H$) for each prototype ($P$). This phenomenon is akin to the mode-collapse problem \cite{thanh2020catastrophic}, well-studied in the literature \cite{zhong2019rethinking}. To mitigate this, we rely on the standard loss of maximizing the entropy of the set of prototypes, making them diverse. Mathematically, it is defined as $\mathcal{L}_{div}(P, H) = -\sum_{i,j} C_{ij}\log(C_{ij})$, where $C = \texttt{Softmax}(PH^T)$.

\textbf{Sparsity}: To further reduce noise in the obtained representation and prevent overfitting, we induce sparsity in the loss function by leveraging the $\ell_1$ norm and $\ell_2$ Frobenius norm regularization of the prototype matrix. Formally $\mathcal{L}_{sp}(P) = \sum_{i,j}\{(P_{ij})^2 + |P_{ij}|\}$.

Finally, the three loss terms above are combined (in $\mathcal{L}_{final}$)
with the task loss ($\mathcal{L}_{task}$) for the downstream task of node classification and node recommendation. $\lambda_{al}$, $\lambda_{d}$, and $\lambda_{s}$ are hyperparameters.

\begin{figure*}[h]
\centering
\resizebox{\textwidth}{!}{%
\fbox{$\displaystyle
\mathcal{L}_{final} = \underbrace{\mathcal{L}_{task}}_{\text{Downstream task loss}} + 
    \underbrace{\lambda_{al} \sum_{l}\mathcal{L}_{alignment} (P_{\mathcal{A}^l})}_{\text{Alignment loss to maximize representation}} +\ 
    \lambda_d \underbrace{\left( \mathcal{L}_{div}(P_{\mathcal{N}}, X) + \sum_l\mathcal{L}_{div}(P_{\mathcal{A}}^l, H^l) \right)}_{\text{Diversity loss to avoid collapse}} +\ 
    \lambda_s \underbrace{\left(\mathcal{L}_{sp}(P_{\mathcal{N}}) + \sum_l\mathcal{L}_{sp}(P_{\mathcal{A}}^l)\right)}_{\text{Sparsity loss to reduce noise}}
$}%
}
\end{figure*}

\begin{table*}[htbp]
  \centering
  \caption{Summary of Research Questions and Findings in the Experimental Evaluation of the Proposed $P^2$GNN Model}
  \label{tab:summary}
  \resizebox{\textwidth}{!}{%
    \begin{tabular}{clc}
    \toprule
    Section & Question & Answer (Yes/No) \\
    \hline
    \multirow{5}{*}{RQ1} & Does the $P^2$GNN model outperform baseline methods on node recommendation tasks? & Yes \\
    & Does the $P^2$GNN model achieve superior performance on node classification tasks? & Yes \\
    & Does the $P^2$GNN model scale effectively to large-scale graphs? & Yes \\
    & Does $P^2$GNN drastically increase training time? & No \\
    & Are the performance improvements of $P^2$GNN statistically significant over the backbone GNN? & Yes \\
    \hline
    \multirow{3}{*}{RQ2} & Does incorporating neighborhood prototypes ($P_\mathcal{N}$) enhance the model's performance? & Yes \\
    & Does incorporating alignment prototypes ($P_\mathcal{A}$) enhance the model's performance? & Yes \\
    & Does the $P^2$GNN model exhibit robustness to noisy node features? & Yes \\
    \hline
    \multirow{3}{*}{RQ3} & Do the learned attention mechanisms contribute to the model's performance improvements? & Yes \\
    & Does $P^2$GNN effectively capture global graph information? & Yes \\
    & Does $P^2$GNN provide effective denoising of node features? & Yes \\
    \hline
\multirow{3}{*}{RQ4} & Can the hyperparameters of the $P^2$GNN model be effectively tuned? & Yes \\
    & Can the optimal number of prototypes be easily determined? & Yes \\
    & Can the loss weights of hyperparameters be tuned to achieve optimal performance? & Yes \\
    \bottomrule
    \end{tabular}%
  }
\end{table*}

\begin{table}[t]
  \centering
  \caption{Results on proprietary e-commerce datasets for the node recommendation task. Results are relative to a production model at an e-commerce store with GCN backbone.}
  \label{tab:prop}
  \small
    \begin{tabular}{lcccccc}
    \toprule
     \textbf{Dataset} & \textbf{Model}  & \textbf{Hits@1} & \textbf{Hits@3} & \textbf{Hits@5} & \textbf{Hits@10} & \textbf{MRR@10} \\
    \midrule
    \multirow{2}{*}{E-comm1} & GCN & 3.77x  & 3.81x  & 3.81x & 3.81x & 3.79x\\
    & $P^2$GNN & \textbf{3.93x} & \textbf{3.95x}  & \textbf{3.96x} & \textbf{3.96x} & \textbf{3.94x}  \\
    \midrule
    \multirow{2}{*}{E-comm2} & GCN & 0.285x  & 0.753x  & 1.056x & 1.531x & 0.613x \\
    & $P^2$GNN & \textbf{0.333x} & \textbf{0.782x}  & \textbf{1.090x} & \textbf{1.568x} & \textbf{0.657x}  \\
    \bottomrule
    \end{tabular}
\end{table}%
% Table generated by Excel2LaTeX from sheet 'Results'
\begin{table*}[tp]
  \centering
%   \tiny
%   \setlength{\tabcolsep}{0.5pt}
  \caption{Average classification accuracy and standard deviation on 8 real-world benchmark datasets with a random 60\%/20\%/20\% data split. The best results are highlighted in \textbf{bold}, and the runner-up score with an underline. NA means that neither the reported results nor the code were available for evaluation. ACM-GCN is used as the backbone for $P^2$GNN across datasets for a fair comparison. In principle, any high-performing GNN can be used as the $P^2$GNN backbone.}
  % \vspace{-3mm}
  \resizebox{\textwidth}{!}{
\begin{tabular}{|c|c|c|c|c|c|c|c|c||c|}
    \hline
    % \toprule
          & \textbf{Cornell} & \textbf{Wisconsin} & \textbf{Texas} & \textbf{Film}  & \textbf{Chameleon} & \textbf{Squirrel} & \textbf{Cora}  & \textbf{CiteSeer}  &  \textbf{Avg. Rank}\\
    % \#nodes & 183   & 251   & 183   & 7,600 & 2,277 & 5,201 & 2,708 & 3,327 & 19,717 &  \\
    % \#edges & 295   & 499   & 309   & 33,544 & 36,101 & 217,073 & 5,429 & 4,732 & 44,338 &  \\
    % \#features & 1,703 & 1,703 & 1,703 & 931   & 2,325 & 2,089 & 1,433 & 3,703 & 500   &  \\
    % \#classes & 5     & 5     & 5     & 5     & 5     & 5     & 7     & 6     & 3     &  \\
    $\mathcal{H}_{node}$ & 0.30 & 0.21 & 0.11 & 0.22 & 0.23 & 0.22 & 0.81 & 0.74  &  \\
    % $H_\text{node}$ & 0.3855 & 0.1498 & 0.0968 & 0.2210 & 0.2470 & 0.2156 & 0.8252 & 0.7175 & 0.7924 &  \\
    % $H_\text{class}$ & 0.0468 & 0.0941 & 0.0013 & 0.0110 & 0.0620 & 0.0254 & 0.7657 & 0.6270 & 0.6641 &  \\
    % Data Splits & 60\%/20\%/20\% & 60\%/20\%/20\% & 60\%/20\%/20\% & 60\%/20\%/20\% & 60\%/20\%/20\% & 60\%/20\%/20\% & 50\%/25\%/25\% & 60\%/20\%/20\% & 60\%/20\%/20\% & 60\%/20\%/20\% &  \\
    % $H_\text{agg}^{M}$ & 0.8032 & 0.7768 & 0.694 & 0.6822 & 0.61  & 0.3566 & 0.9904 & 0.9826 & 0.9432 &  \\
\hline
    % \midrule
        %   & \multicolumn{9}{c|}{Test Accuracy (\%) of State-of-the-art Models, Baseline GNN Models and ACM-GNN models} & Rank \\
    % \midrule
    MLP & 91.30 $\pm$ 0.70 & {93.87 $\pm$ 3.33} & 92.26 $\pm$ 0.71 & 38.58 $\pm$ 0.25 & 46.72 $\pm$ 0.46 & 31.28 $\pm$ 0.27 & 76.44 $\pm$ 0.30 & 76.25 $\pm$ 0.28 & 11.0 \\
    % \midrule
    GAT~\cite{velivckovic2017attention}   & 76.00 $\pm$ 1.01 & 71.01 $\pm$ 4.66 & 78.87 $\pm$ 0.86 & 35.98 $\pm$ 0.23 & 63.9 $\pm$ 0.46 & 42.72 $\pm$ 0.33 & 76.70 $\pm$ 0.42 & 67.20 $\pm$ 0.46  & 13.6 \\
    % APPNP~\cite{klicpera2018predict} & 91.80 $\pm$ 0.63 & 92.00 $\pm$ 3.59 & 91.18 $\pm$ 0.70 & 38.86 $\pm$ 0.24 & 51.91 $\pm$ 0.56 & 34.77 $\pm$ 0.34 & 79.41 $\pm$ 0.38 & 68.59 $\pm$ 0.30 & 11.25 \\
    GPRGNN~\cite{chien2021adaptive} & 91.36 $\pm$ 0.70 & 93.75 $\pm$ 2.37 & 92.92 $\pm$ 0.61 & 39.30 $\pm$ 0.27 & 67.48 $\pm$ 0.40 & 49.93 $\pm$ 0.53 & 79.51$\pm$ 0.36 & 67.63 $\pm$ 0.38 & 8.4 \\
    H2GCN~\cite{zhu2020beyond} & 86.23 $\pm$ 4.71 & 87.5 $\pm$ 1.77 & 85.90 $\pm$ 3.53 & 38.85 $\pm$ 1.17 & 52.30 $\pm$ 0.48 & 30.39 $\pm$ 1.22 & 87.52 $\pm$ 0.61 & 79.97 $\pm$ 0.69 & 10.8 \\
    % MixHop~\cite{abu2019mixhop} & 60.33 $\pm$ 28.53 & 77.25 $\pm$ 7.80 & 76.39 $\pm$ 7.66 & 33.13 $\pm$ 2.40 & 36.28 $\pm$ 10.22 & 24.55 $\pm$ 2.60 & 65.65 $\pm$ 11.31 & 49.52 $\pm$ 13.35 &  18 \\
    GCN+JK~\cite{kipf2016classification} & 66.56 $\pm$ 13.82 & 62.50 $\pm$ 15.75 & 80.66 $\pm$ 1.91 & 32.72 $\pm$ 2.62 & 64.68 $\pm$ 2.85 & {53.40 $\pm$ 1.90} & 86.90 $\pm$ 1.51 & 73.77 $\pm$ 1.85 &  13.1 \\
     GAT+JK~\cite{velivckovic2017attention} & 74.43 $\pm$ 10.24 & 69.50 $\pm$ 3.12 & 75.41 $\pm$ 7.18 & 35.41 $\pm$ 0.97 & 68.14 $\pm$ 1.18 & 52.28 $\pm$ 3.61 & 89.52 $\pm$ 0.43 & 74.49 $\pm$ 2.76 &  11.0 \\
    FAGCN~\cite{bo2021beyond} & 88.03 $\pm$ 5.6 & 89.75 $\pm$ 6.37 & 88.85 $\pm$ 4.39 & 31.59 $\pm$ 1.37 & 49.47 $\pm$ 2.84 & 42.24 $\pm$ 1.2 & 88.85 $\pm$ 1.36 & \textbf{82.37 $\pm$ 1.46} &  9.1 \\
    BernNet~\cite{he2021bernnet} & 92.13 $\pm$ 1.64 & NA    & 93.12 $\pm$ 0.65 & {41.79 $\pm$ 1.01} & {68.29 $\pm$ 1.58} & 51.35 $\pm$ 0.73  & 88.52 $\pm$ 0.95 & 80.09 $\pm$ 0.79 & 7.5 \\
    % GraphSAGE~\cite{hamilton2017inductive} & 71.41 $\pm$ 1.24 & 64.85 $\pm$ 5.14 & 79.03 $\pm$ 1.20 & 36.37 $\pm$ 0.21 & 62.15 $\pm$ 0.42 & 41.26 $\pm$ 0.26 & 86.58 $\pm$ 0.26 & 78.24 $\pm$ 0.30 & 14.125 \\
    % Geom-GCN & 60.81 & 64.12 & 67.57 & 31.63 & 60.9  & 38.14    & 85.27 & 77.99 & 27.44 \\
    % \midrule
    % SGC-1~\cite{wu2019simplifying} & 70.98 $\pm$ 8.39 & 70.38 $\pm$ 2.85 & 83.28 $\pm$ 5.43 & 25.26 $\pm$ 1.18 & 64.86 $\pm$ 1.81 & 47.62 $\pm$ 1.27 & 85.12 $\pm$ 1.64 & 79.66 $\pm$ 0.75 & 85.5 $\pm$ 0.76 & 24.90 \\
    SGC-2~\cite{wu2019simplifying} & 72.62 $\pm$ 9.92 & 74.75 $\pm$ 2.89 & 81.31 $\pm$ 3.3 & 28.81 $\pm$ 1.11 & 62.67 $\pm$ 2.41 & 41.25 $\pm$ 1.4 & 85.48 $\pm$ 1.48 & 80.75 $\pm$ 1.15  & 13.3 \\
    % GCNII* & 90.49 $\pm$ 4.45 & 89.12 $\pm$ 3.06 & 88.52 $\pm$ 3.02 & 41.54 $\pm$ 0.99 & 62.8 $\pm$ 2.87 & 38.31 $\pm$ 1.3 & 88.93 $\pm$ 1.37 & 81.83 $\pm$ 1.78 & 89.98 $\pm$ 0.52 & 16.40 \\
    % GCN~\cite{kipf2016classification}   & 82.46 $\pm$ 3.11 & 75.5 $\pm$ 2.92 & 83.11 $\pm$ 3.2 & 35.51 $\pm$ 0.99 & 64.18 $\pm$ 2.62 & 44.76 $\pm$ 1.39 & 87.78 $\pm$ 0.96 & 81.39 $\pm$ 1.23 & 88.9 $\pm$ 0.32 & 20.90 \\
    Snowball-2~\cite{luan2019break} & 82.62 $\pm$ 2.34 & 74.88 $\pm$ 3.42 & 83.11 $\pm$ 3.2 & 35.97 $\pm$ 0.66 & 64.99 $\pm$ 2.39 & 47.88 $\pm$ 1.23 & 88.64 $\pm$ 1.15 & 81.53 $\pm$ 1.71 & 9.8 \\
    Snowball-3~\cite{luan2019break} & 82.95 $\pm$ 2.1 & 69.5 $\pm$ 5.01 & 83.11 $\pm$ 3.2 & 36.00 $\pm$ 1.36 & 65.49 $\pm$ 1.64 & 48.25 $\pm$ 0.94 & \underline{89.33 $\pm$ 1.3} & 80.93 $\pm$ 1.32 & 9.3 \\
    
    % \midrule
    GloGNN++~\cite{li2022finding} & 81.96 $\pm$ 5.1 & 83.75 $\pm$ 3.53 & 85.24 $\pm$ 4.9 & 38.8 $\pm$ 1.3 & \textbf{71.33 $\pm$ 2.42} & 42.65 $\pm$ 1.76 & 74.87 $\pm$ 1.09 & 76.53 $\pm$ 1.65 & 10.3\\
    GCNII~\cite{chen2020simple} & 89.18 $\pm$ 3.96 & 83.25 $\pm$ 2.69 & 82.46 $\pm$ 4.58 & 40.82 $\pm$ 1.79 & 60.35 $\pm$ 2.7 & 38.81 $\pm$ 1.97 & 88.98 $\pm$ 1.33 & 81.58 $\pm$ 1.3 & 9.1 \\
    GGCN~\cite{yan2021two} &  92.30 $\pm$ 3.67 & 87.25 $\pm$ 4.39 & 90.16 $\pm$ 2.48 & 39.29 $\pm$ 0.48 & 69.5 $\pm$ 2.11 & 55.67 $\pm$ 1.87 & 88.57 $\pm$ 1.11 & 78.2 $\pm$ 0.68 & 6.4\\ 
    % ACM-GCN~\cite{luan2022revisiting} & \underline{93.77 $\pm$ 1.91} & 93.25 $\pm$ 2.92 & \underline{93.61 $\pm$ 1.55} & 39.33 $\pm$ 1.25 & 63.68 $\pm$ 1.62 & 46.4 $\pm$ 1.13 & 86.63 $\pm$ 1.13 & 80.96 $\pm$ 0.93 & 6.375 \\
    ACM-GCN~\cite{luan2022revisiting} & \underline{94.43 $\pm$ 2.76} & \underline{96.12 $\pm$ 2.05} & \underline{95.57 $\pm$ 3.60} & {41.98 $\pm$ 0.89} & 68.45 $\pm$ 1.67 & 56.31 $\pm$ 1.23 & 88.56 $\pm$ 0.87 & \underline{81.96 $\pm$ 1.48} & \underline{3.6}\\
    VR-GNN~\cite{vrgnn} & 92.70 $\pm$ 2.70 & NA & 94.86 $\pm$ 1.89 & \underline{42.16 $\pm$ 0.42} & \underline{71.21 $\pm$ 1.17} & \textbf{57.50 $\pm$ 1.18} & 88.27 $\pm$ 0.89 & 81.95 $\pm$ 0.77 & 5.0\\
    \hline
    $P^2$GNN & \textbf{95.41 $\pm$ 2.72} & \textbf{98.00 $\pm$ 1.50} & \textbf{96.72 $\pm$ 1.47} & \textbf{43.05 $\pm$ 1.34} & {69.87 $\pm$ 1.63} & \underline{56.47 $\pm$ 1.36} & \textbf{89.89 $\pm$ 0.90} & \textbf{82.37 $\pm$ 1.09} & \textbf{1.5}\\
    \hline
    % \hline
          {$p$-value} & 0.039 & 0.00195 & 0.04326 & 0.00803 & 0.0179 & 0.3575 & 0.00006 & 0.2948 & - \\
    \hline
    % \bottomrule
    \end{tabular}%
  \label{tab:ours_vs_sota_random}%
  }
\end{table*}%

\section{Experiments}
In this section, we comprehensively evaluate and investigate the efficacy of our approach to answer the following research questions. We summarize the key findings and conclusions of each research experiment in Table~\ref{tab:summary}.

\begin{itemize}[noitemsep,topsep=0pt,leftmargin=*]
    \item \textbf{RQ1}: How does $P^2$GNN perform against state-of-the-art (SOTA) models on different datasets? 
    \item \textbf{RQ2}: How does each prototype of the $P^2$GNN contribute to performance improvement? Does the model boost performance on different node segments differently?
    \item \textbf{RQ3}: How much do the prototypes contribute to improvements, and is the desired objective of each prototype achieved?
    \item \textbf{RQ4}: How sensitive is $P^2$GNN w.r.t. different hyperparameters?
\end{itemize}

{\bf Datasets and Experimental Settings:} For fair evaluation, we conducted experiments on 16 commonly used open-source and 2 proprietary e-commerce datasets spanning various domains, scales, and graph heterophilies. We performed node classification on open-source and node recommendation on proprietary datasets. Table~\ref{tab:data_stats} in Appendix~\ref{sec:datasets} summarizes the dataset details. To mitigate noise impact, we repeated experiments 10 times on 8 small-scale \cite{pei2020geom, musae} and 5 times on 6 large-scale \cite{lim2021large, lim2021new} datasets, reporting average test accuracy and standard deviation. For the proprietary and fraud datasets, the experiments were repeated once. Additional implementation details are provided in Appendix~\ref{sec:implementaion}.

{\bf More details on Proprietary E-commerce Datasets:} We evaluate our approach on two proprietary node recommendation datasets, E-comm1 and E-comm2, currently utilized in production at an e-commerce store. These unweighted and directed graph datasets, ranging from 700K to 3.5M nodes and 10M to 400M edges, represent products as nodes and co-purchase or co-view relationships as directed edges. While we cannot disclose exact details due to confidentiality, these datasets encompass diverse structural aspects. In production, GCN models are used to train on these datasets, and we benchmark our $P^2$GNN technique by integrating it with the deployed GCN model. We assess its effectiveness in practical scenarios, robustness across different graph structures and sizes, and applicability to graph-based applications.

{\bf Baselines:}
We compare our method with the following baselines: (1) MLP; (2) General GNN methods: GCN~\cite{kipf2016classification}, GAT~\cite{velivckovic2017attention}, MixHop~\cite{abu2019mixhop}, and GCNII~\cite{chen2020simple}; (3) Heterophilous graph-oriented methods: H2GCN~\cite{zhu2020beyond}, GPR-GNN~\cite{chien2021adaptive}, GloGNN++~\cite{li2022finding}, and ACM-GCN~\cite{luan2022revisiting}. Additional SOTA baselines are included in the respective tables for different dataset types. Following prior works~\cite{luan2022revisiting}, we report the baseline results as mentioned in the original papers\footnote{If the results are unavailable, we run the code provided by the authors. When the code is not available, we report NA, following standard practice.}.

\begin{figure*}[tp]
  \centering
  \subfigure{\includegraphics[width=0.24\textwidth]{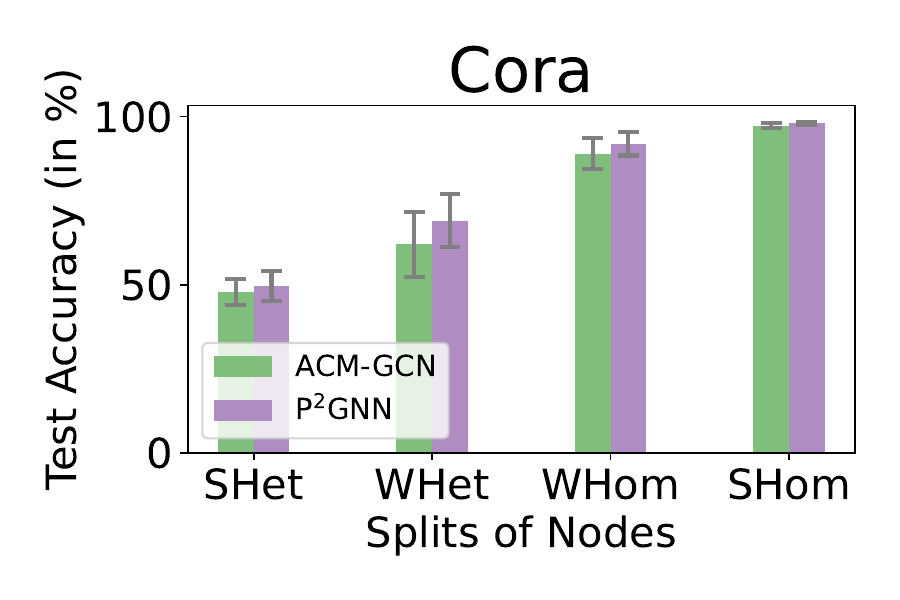}}
\subfigure{\includegraphics[width=0.24\textwidth]{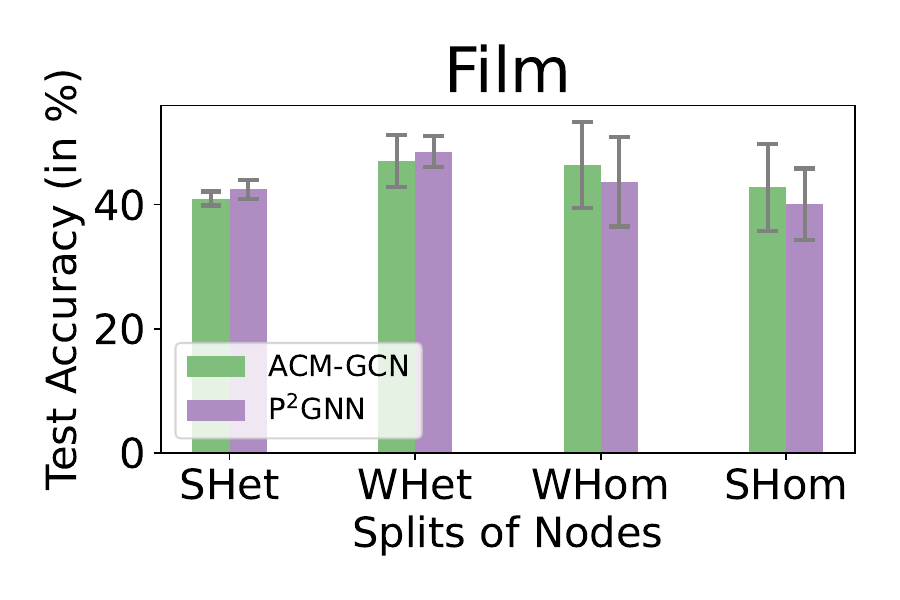}}
\subfigure{\includegraphics[width=0.24\textwidth]{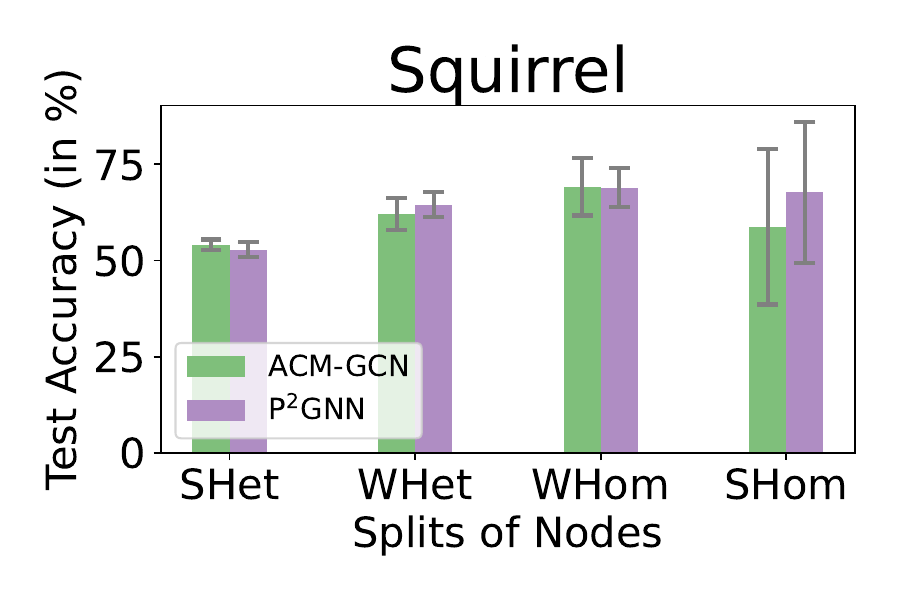}}
\subfigure{\includegraphics[width=0.24\textwidth]{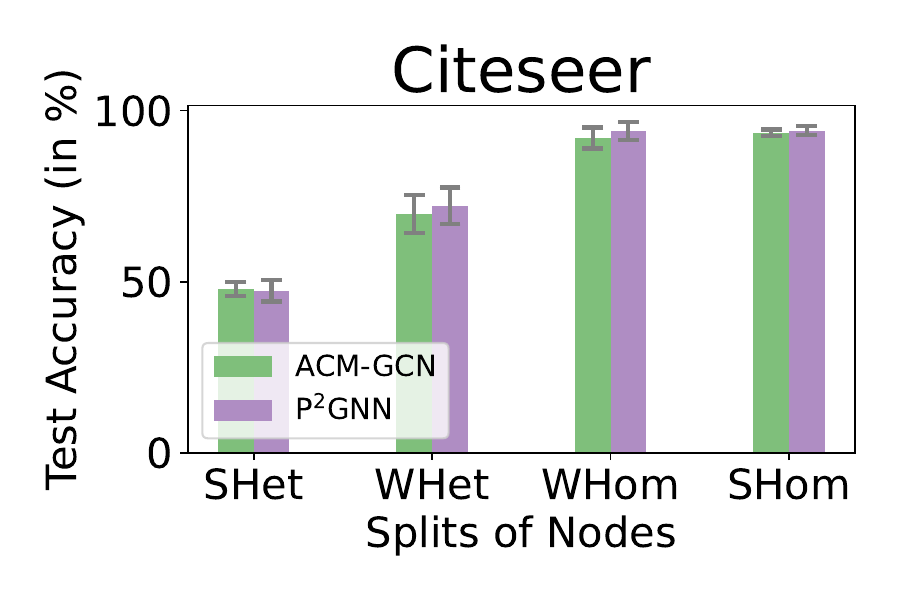}}
  \caption{Test accuracy performance of $P^2$GNN compared against its backbone ACM-GCN, segmented by the level of heterophily in test-nodes. $P^2$GNN consistently boosts the performance on strong (SHet) and weak (WHet) heterophilous node segments.}
  \label{fig:sw_homo_hetero}
\end{figure*}

\begin{table*}
	\begin{minipage}{0.25\linewidth}
		\caption{Results (\%) on 2 fraud detection datasets. We compare the AUC-ROC score for this task. The best score on each dataset is bolded and the runner-up is underlined.}
  \label{tab:fraud}
  \resizebox{\linewidth}{!}
  {
    \begin{tabular}{|c||c|c|c|c|c|c|c|c|}
    \hline
     Model  & \textbf{YelpChi} & \textbf{Amazon}\\
    \hline
    MLP & 73.66 & 90.82 \\
    GCN~\cite{kipf2016classification} & 54.13 & 50.83 \\
    GAT~\cite{velivckovic2017attention} & 54.59 & 77.32 \\
    JKNet~\cite{xu2018representation} & 77.36 & 89.70 \\
    GPRGNN~\cite{chien2020adaptive} & \underline{83.55} & 93.58 \\
    CARE-GNN~\cite{dou2020enhancing} & 73.00 & 88.32 \\
    PC-GNN~\cite{liu2021pick} & 81.18 & 95.55 \\
    BWGNN~\cite{tang2022rethinking} & 82.55 & 93.95 \\
    GHRN~\cite{gao2023addressing}  & 82.88 &\underline{96.19}\\
    \hline
    $P^2$GNN & \textbf{84.43} & \textbf{97.64}\\
    \hline
    \end{tabular}%
    }
	\end{minipage}
	\hfill
	\begin{minipage}{0.45\linewidth}
		\centering
		\subfigure{\includegraphics[width=0.48\textwidth]{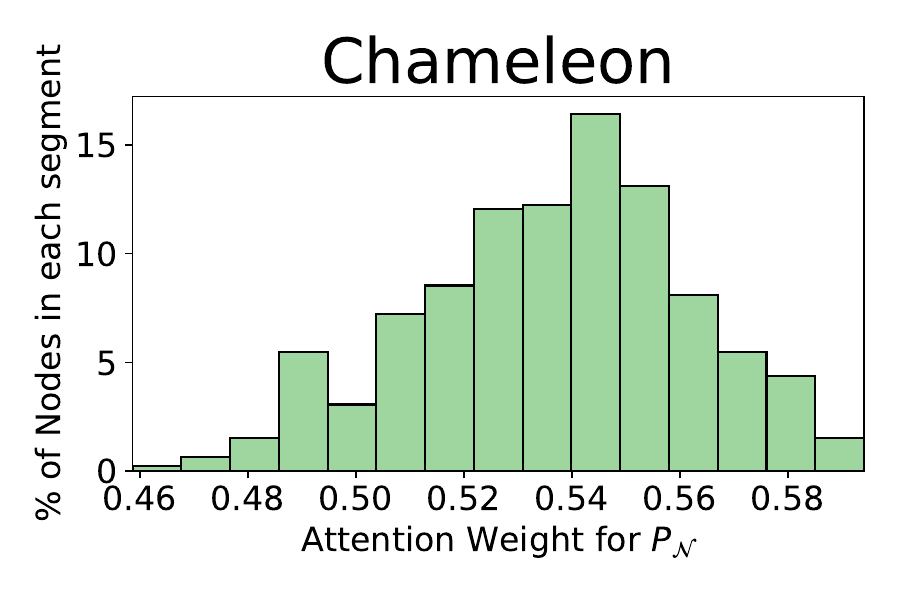}}
    \subfigure{\includegraphics[width=0.48\textwidth]{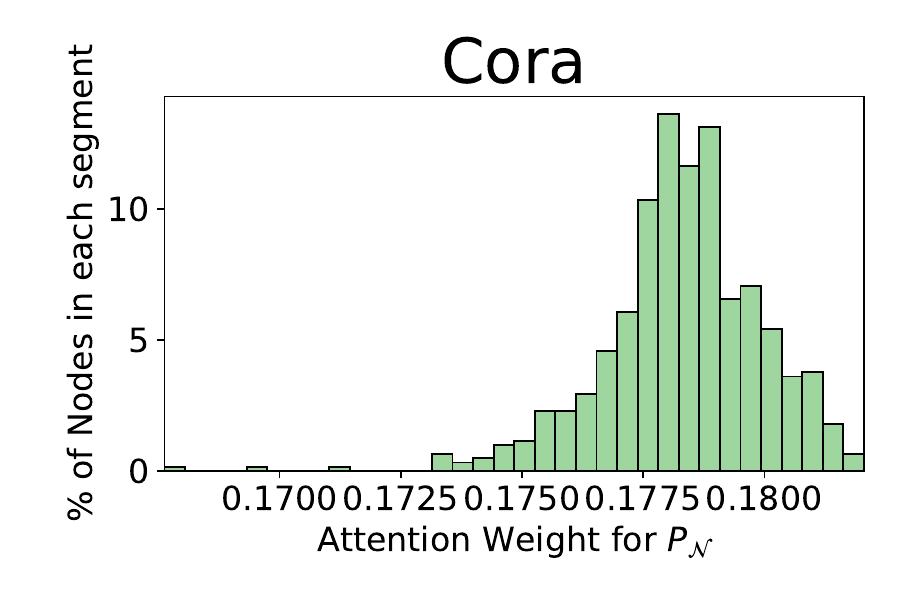}}
    \subfigure{\includegraphics[width=0.48\textwidth]{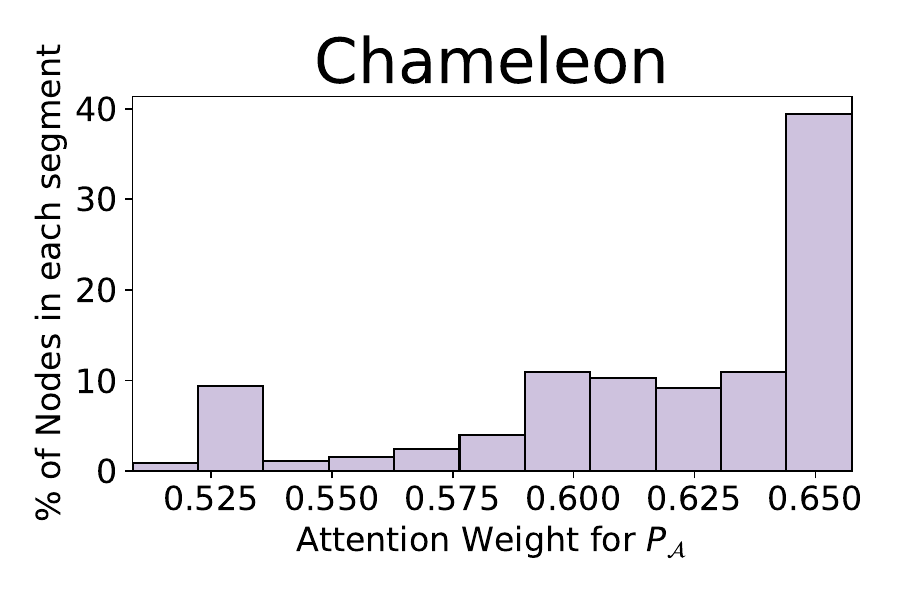}}
\subfigure{\includegraphics[width=0.48\textwidth]{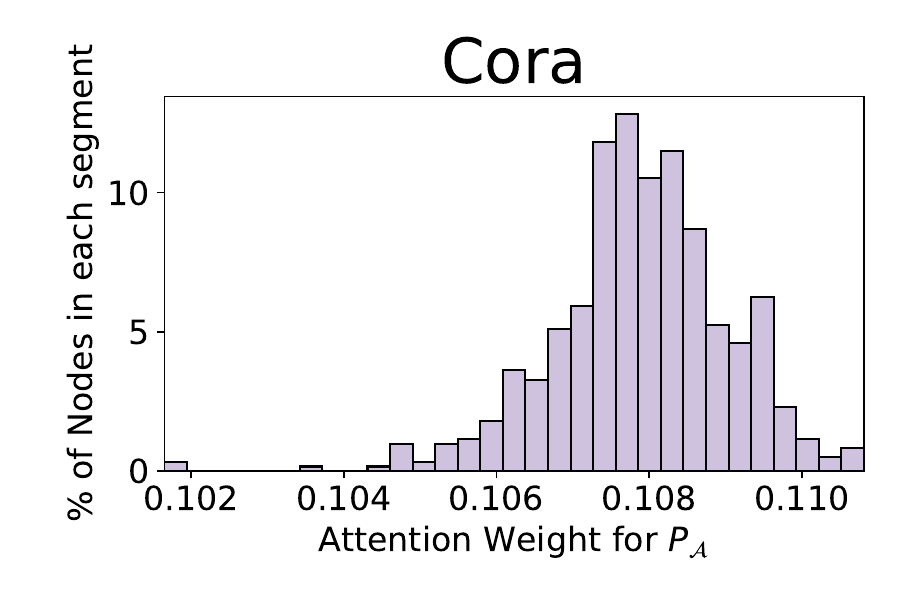}}
  \captionof{figure}{The plots illustrate the level of attention given by test nodes to two types of prototypes. The top row depicts the attention paid to the prototypes $P_{\mathcal{N}}$, while the bottom row shows the attention given to prototypes $P_{\mathcal{A}}$.}
  \label{fig:attention-plots}
	\end{minipage}
	\hfill
	\begin{minipage}{0.25\linewidth}
		\centering
		\subfigure{\includegraphics[width=0.96\textwidth]{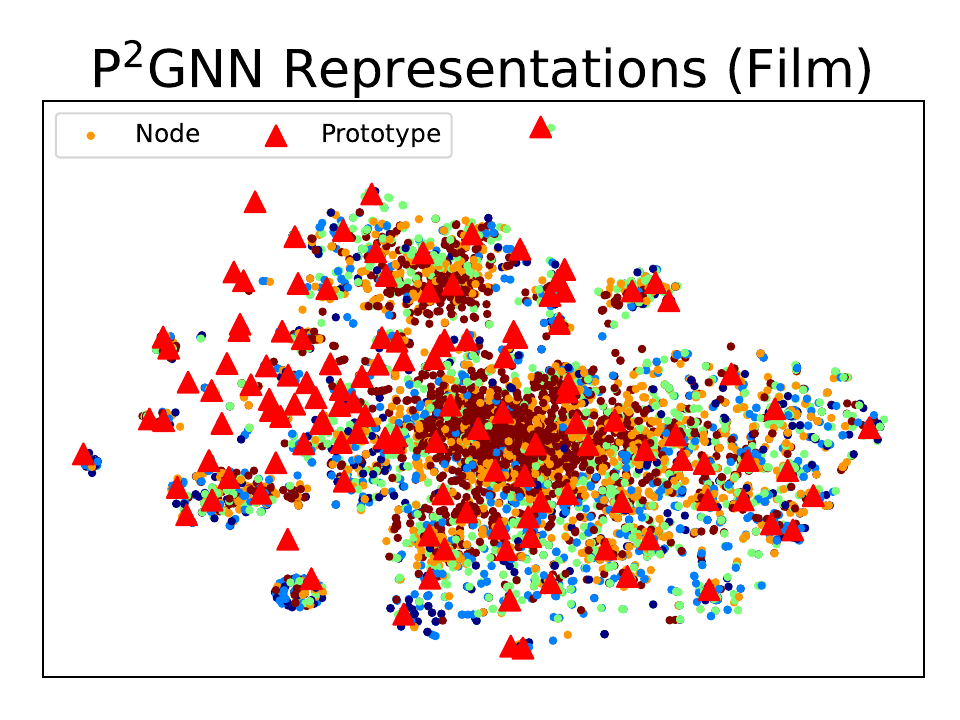}}
\subfigure{\includegraphics[width=0.96\textwidth]{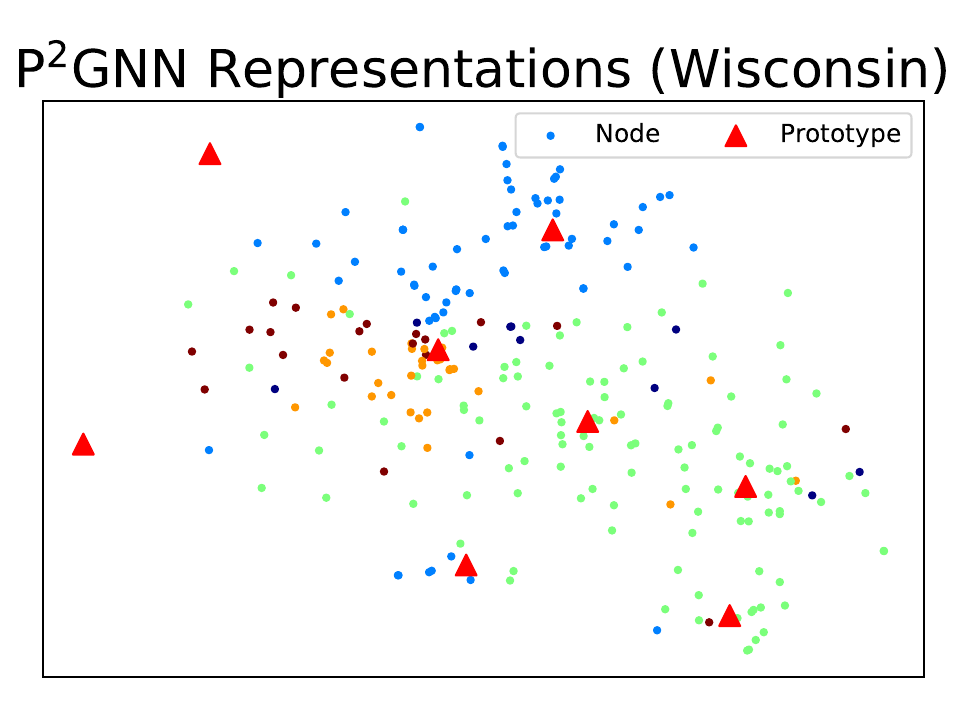}}
  \captionof{figure}{Learned Prototype $P_{\mathcal{N}}$ representation for first GNN layer and raw node features.}
  \label{fig:p2gnn-representation}
	\end{minipage}
\end{table*}

\subsection{RQ1: Performance Comparison}
We evaluate $P^2$GNN on several key aspects below:

\textbf{Node Recommendation Task:} The results presented in Table~\ref{tab:prop} demonstrate the superior performance of the proposed $P^2$GNN model in enhancing node recommendation on proprietary datasets, surpassing the existing GNN models with GCN backbone currently in production at an e-commerce store. While absolute values cannot be disclosed due to confidentiality reasons, the relative improvements are noteworthy. On the E-comm1 dataset, $P^2$GNN achieves an impressive average relative improvement of 4\% across various evaluation metrics, while on the E-comm2 dataset, it delivers a notable 1.5\% performance gain. These substantial improvements in recommendation quality directly translate into tangible benefits for users, aligning with the key objectives of the recommendation system. \textit{Key takeaway: $P^2$GNN outperforms existing GNN models deployed in production for node recommendation.} 

\textbf{Node Classification Performance:} Table~\ref{tab:ours_vs_sota_random} compares $P^2$GNN with baselines on open-source graph datasets. $P^2$GNN demonstrates consistent performance gains across diverse datasets, ranging from high homophily (low noise) to low homophily (high noise) over the ACM-GCN backbone. We achieved SOTA results on the majority of datasets, except for Squirrel and Chameleon. This is due to the presence of a large number of duplicate nodes that leads to train-test data leakage ~\cite{platonov2023a}. However, we were still able to outperform its backbone ACM-GCN on these datasets. Table~\ref{tab:fraud} highlights the applicability and efficacy of $P^2$GNN on two fraud detection tasks, showcasing its versatility. Additional experiments on open-source datasets, including a comparison with recent prototype-based techniques, are presented in Appendix A. \textit{Key takeaway: $P^2$GNN consistently outperforms baselines across various open-source datasets, demonstrating its effectiveness in diverse settings.}

\begin{table*}[h]
\centering
%\resizebox{0.87\linewidth}{!}{
\caption{Results (\%) on 6 large-scale datasets released in~\cite{lim2021large}.
We compare the AUC score on 
genius as in~\cite{lim2021large}.
For other datasets, 
we show the classification accuracy.
The error bar ($\pm$) denotes the standard deviation score of results over 5 trials. 
We highlight the best score on each dataset in bold and the runner-up score with an underline.
Note that approaches such as \cite{maurya2022not, zhu2020beyond, suresh2021breaking} suffers out-of-memory issue; We choose not to include them on large datasets.}
% \vspace{-3mm}
\label{tab:ours_vs_sota_large}
\resizebox{0.7\linewidth}{!}
{
\begin{tabular}{|c|c|c|c|c|c|c||c|}
\hline
                    & \textbf{Penn94} & \textbf{pokec}     & \textbf{arXiv-year} & \textbf{snap-patents} & \textbf{genius} & \textbf{twitch-gamers} & \textbf{Avg. Rank}
                    % \parbox[t]{2mm}{\multirow{5}{*}{\rotatebox[origin=c]{90}{\textbf{Avg. Rank}}}} 
                    \\
$\mathcal{H}_{node}$  &  0.47     &    0.44       &  0.22  & 0.07   & 0.61  &  0.54 &  \\
% \textbf{\#Nodes}  &   41,554    &      1,632,803     & 169,343   &  2,923,922 &  421,961 & 168,114  &  \\
%  \textbf{\#Edges}  &  1,362,229     &    30,622,564       &  1,166,243  &  13,975,788  & 984,979  &  6,797,557 &   \\
%   \textbf{\#Features}  & 5     &     65      &    128 &  269   & 12  &  7 &   \\
%  \textbf{\#Classes} &    2   &     2      & 5  &  5  & 2 & 2  & \\ 
\hline   
     MLP                   &  $ 73.61 \pm 0.40 $  &     $ 62.37 \pm 0.02$        &  $ 36.70 \pm 0.21 $  &  $ 31.34 \pm 0.05 $  &  $ 86.68 \pm 0.09 $ & $ 60.92 \pm 0.07 $  & 10.5\\
     GCN                     &  $ 82.47 \pm 0.27 $    &  $ 75.45 \pm 0.17 $          & $ 46.02 \pm 0.26 $ & $45.65 \pm 0.04 $ &   $87.42 \pm 0.37 $  & $ 62.18 \pm 0.26 $ & 7.8\\
     GAT                    &  $ 81.53  \pm 0.55 $   &   $71.77  \pm 6.18 $          &$ 46.05 \pm 0.51 $ & $ 45.37 \pm 0.44 $ &  $ 55.80 \pm 0.87 $ &  $ 59.89 \pm 4.12$  &  9.3 \\    
      MixHop                &   $ 83.47 \pm 0.71 $   &     $ 81.07 \pm 0.16 $       & $ 51.81 \pm 0.17$ & $ 52.16 \pm 0.09 $ & $90.58 \pm 0.16 $ & $ 65.64 \pm 0.27$  &  5.2 \\  
%     Geom-GCN                  &  $ \pm $   &    $ \pm $         &$\pm $ & $\pm $ & $\pm $ &  $ \pm $  & \\     
     GCN\rom{2}                  & $ 82.92 \pm 0.59 $  &   $78.94  \pm 0.11$          & $ 47.21\pm 0.28$ &  $ 37.88 \pm 0.69$  & $90.24 \pm 0.09$ & $ 63.39 \pm 0.61$  & 6.8 \\
%   H$_2$GCN               &   $81.31\pm 0.60$    &    OOM       & $49.09 \pm 0.10$  &  OOM  &  OOM  & OOM   &  10.83 \\ 
%   WRGAT               &     $74.32\pm0.53$   &   OOM          &  OOM &  OOM  & OOM  & OOM  &  12.83  \\ 
    GPR-GNN              &   $81.38 \pm 0.16 $   &     $ 78.83 \pm 0.05 $       & $ 45.07 \pm 0.21$ & $ 40.19 \pm 0.03 $ & $ 90.05 \pm 0.31$ & $61.89 \pm 0.29$  &  8.7 \\ 
    % GGCN               &    OOM    &     OOM        & OOM &   OOM & OOM & OOM  &   12.25  \\ 
    % Dual-Net GNN & $85.81 \pm 0.48$ & OOM & OOM & OOM & $90.79 \pm 0.81$ & OOM \\
    ACM-GCN               &    $82.52\pm 0.96$    &    $63.81\pm5.20$         & $47.37 \pm 0.59$ &  $55.14\pm0.16$  &  $80.33 \pm 3.91$ &  $62.01\pm 0.73$&   7.7 \\ 
       LINKX                &   $ 84.71 \pm 0.52 $   &     $ 82.04  \pm 0.07 $       & $ \bm{56.00 \pm 1.34}$ & $ 61.95 \pm 0.12$ & $  90.77 \pm 0.27 $ & $66.06 \pm 0.19$  &   3.3  \\ 
     GloGNN               &   $ 85.57  \pm 0.35$   &  $83.00 \pm 0.10$       & $ 54.68 \pm 0.34$ & $ 62.09 \pm 0.27$ & $ 90.66 \pm 0.11$ & $ 66.19  \pm 0.29$  &  3.2 \\  
      GloGNN++               &   $  \underline{85.74  \pm 0.42} $   &  $  \underline{83.05\pm 0.07}$       & $ 54.79  \pm 0.25$ & $  \underline{62.03 \pm 0.21} $ & $ \underline{90.91 \pm 0.13}$ &  \underline{$66.34 \pm 0.29$}  &  \underline{2.3}   \\  
    %   ProtoGNN               &   $ 86.23  \pm 0.34$   &  $81.94 \pm 0.10$       & xx & xx & $ 91.18 \pm 0.13$ & $ 66.42  \pm 0.34$  &  - \\  
      \hline
      $P^2$GNN & \bm{$86.40 \pm 0.55$} & \bm{$83.30 \pm 0.2$} & \underline{$54.88 \pm 0.34$} & \bm{$62.73 \pm 0.07$} & \bm{$91.01 \pm 0.10$} & \bm{$66.55 \pm 0.16$} & \bm{$1.2$} \\
      \hline
      % \vspace{-3mm}
\end{tabular}

}
\end{table*}

% Table generated by Excel2LaTeX from sheet 'Results'
\begin{table}[htbp]
  \centering
  % \aboverulesep = 0pt
  % \belowrulesep = 0pt
%   \tiny
  \caption{Comparison of average running time per epoch (in seconds) of $P^2$GNN with ACM-GNN on large datasets}
  % \vspace{-3mm}
  \label{tab:time_study}
  \resizebox{\linewidth}{!}
  {
    \begin{tabular}{|c||c|c|c|c|c|c|c|c|}
    \hline
     Model  & \textbf{Penn94} & \textbf{pokec}     & \textbf{arXiv-year} & \textbf{snap-patents} & \textbf{genius} & \textbf{twitch-gamers} \\
    \hline
    ACM-GNN & 1.52  & 2.51  & 1.02 & 3.43& 1.19  & 1.23 \\
    $P^2$GNN & 2.02 & 2.92  & 1.32 & 3.72  & 1.46  & 1.84  \\
        %   & $\checkmark$ & $\checkmark$ & 5.24  & 5.27  & 5.46  & 5.72  & 5.65  & 7.87  & 5.48  & 5.65 \\
          % & $\checkmark$ & $\checkmark$ & $\checkmark$ & 7.59  & 8.28  & 8.06  & 8.85  & 8     & 10    & 8.27  & 8.5   & 8.68  &  \\
    \hline
    % \bottomrule
    \end{tabular}%
    }
    % \vspace{-5mm}
\end{table}%

\textbf{Performance on Large datasets:} $P^2$GNN improves the accuracy of SOTA GNN models like GloGNN++~\cite{li2022finding} and ACM-GCN~\cite{luan2022revisiting} on large-scale datasets, establishing itself as the highest-ranking approach as shown in Table~\ref{tab:ours_vs_sota_large}. Table~\ref{tab:time_study} draws out the fact that average running time per-epoch for $P^2$GNN + ACM-GCN is largely comparable with the backbone ACM-GCN model. \textit{Key takeaway: $P^2$GNN achieves superior performance on large-scale datasets while maintaining computational efficiency.}

\textbf{Statistical Significance:} Table~\ref{tab:ours_vs_sota_random} also showcases the statistical significance of the improvement by $P^2$GNN against its backbone ACM-GCN. The significance is measured by a paired t-test on 40 runs of the two models with different seeds. We observe a consistent (8/8 datasets) improvement in mean test accuracy, with the difference being significant for 6 out of 8 datasets (p-value < 0.05), showing that our improvements are statistically significant over and above the baseline models. \textit{Key takeaway: The performance gains of $P^2$GNN over the baseline ACM-GCN are statistically significant, demonstrating the robustness of the proposed approach.}

\subsection{RQ2: Ablation Study}
$P^2$GNN is a generic and extensible framework that allows incorporating the concept of two types of prototypes into any existing GNN layer as a backbone. We investigate the individual contributions of the two prototype methods introduced in Sections~\ref{sec:pan} and~\ref{sec:pma} by integrating them into three different backbones: GCN~\cite{kipf2016classification}, SGC~\cite{wu2019simplifying}, and ACM-GNN~\cite{luan2022revisiting}, and evaluating their performance on small open-source datasets. As shown in Table~\ref{tab:ablation_study}, the addition of prototypes to any of the GNN layer backbones improves their performance. This observation holds true for either prototype combination, validating that incorporating prototypes as neighbors and alignment has the potential to enhance model performance, regardless of the underlying GNN backbone layers. \textit{Key takeaway: Incorporating prototypes as neighbors and alignment improves the performance of various GNN backbones.}
% Table generated by Excel2LaTeX from sheet 'Results'
\begin{table*}[htbp]
  % \renewcommand{\arraystretch}{2}%
%   \vspace{-7mm}
  \centering
  % \aboverulesep = 0pt
  % \belowrulesep = 0pt
%   \tiny
  \caption{Ablation study  of $P^2$GNN on 8 real-world datasets using random 60\%/20\%/20\% data split. The "Models" refers to backbone chosen for $P^2$GNN addition with \cmark meaning the component is added to backbone. The best test results are bolded.}
  % \vspace{-3mm}
  \label{tab:ablation_study}
  \resizebox{\linewidth}{!}
  {
    \begin{tabular}{|c|cc|c|c|c|c|c|c|c|c|}
    \hline
    % \toprule
    % \multicolumn{13}{c}{Ablation Study on Different Components in GCN, SGC and ACM-GCN (in \%)}  \\
    % \midrule  
     & \multicolumn{2}{c|}{Model Components} & \textbf{Cornell} & \textbf{Wisconsin} & \textbf{Texas} & \textbf{Film}  & \textbf{Chameleon} & \textbf{Squirrel} & \textbf{Cora}  & \textbf{CiteSeer} \\
  % \cmidrule{2-12}    
    Models & $P_{\mathcal{N}}$    & $P_{\mathcal{A}}$  & Acc $\pm$ Std & Acc $\pm$ Std & Acc $\pm$ Std & Acc $\pm$ Std & Acc $\pm$ Std & Acc $\pm$ Std & Acc $\pm$ Std & Acc $\pm$ Std \\
    \hline
    \multicolumn{1}{|c|}{\multirow{3}[1]{*}{GCN w/}} & \xmark&   \xmark    & 82.30 $\pm$ 2.82 & 72.50 $\pm$ 3.71 & 82.62 $\pm$ 4.82 & 35.52 $\pm$ 0.77 & 64.38 $\pm$ 2.44 & 45.25 $\pm$ 2.07 & 87.88 $\pm$ 0.81 & 81.39 $\pm$ 1.21\\
      & \cmark &  \xmark       & 84.26 $\pm$ 3.45 & 78.25 $\pm$ 4.75 & 83.61 $\pm$ 3.20 & 36.45 $\pm$ 1.29 & 64.49 $\pm$ 2.56 & 45.37 $\pm$ 1.76 & 89.16 $\pm$ 1.01 & 80.46 $\pm$ 2.12 \\
      & \cmark &   \cmark    &   84.10 $\pm$ 2.84 & 78.38 $\pm$ 4.61 & 83.44 $\pm$ 3.47 & 36.81 $\pm$ 0.94 & 64.22 $\pm$ 2.37 & 45.37 $\pm$ 1.37 & 89.97 $\pm$ 0.89 & 80.72 $\pm$ 1.57\\
    \hline
    \multicolumn{1}{|c|}{\multirow{3}[0]{*}{SGC w/}} & \xmark &   \xmark          & 70.98 $\pm$ 8.39 & 70.38 $\pm$ 2.85 & 83.28 $\pm$ 5.43 & 25.26 $\pm$ 1.18 & 51.53 $\pm$ 1.52 & 31.89 $\pm$ 1.85 & 76.60 $\pm$ 2.62 & 79.02 $\pm$ 1.74  \\
          & \cmark & \xmark &        78.20 $\pm$ 4.02 & 72.75 $\pm$ 4.99 & 83.93 $\pm$ 2.72 & 32.63 $\pm$ 2.24 & 53.81 $\pm$ 1.70 & 37.69 $\pm$ 1.58 & 80.11 $\pm$ 2.87 & 78.99 $\pm$ 1.60  \\
          & \cmark & \cmark  &   80.98 $\pm$ 3.30 & 74.50 $\pm$ 4.12 & 84.10 $\pm$ 3.81 & 33.26 $\pm$ 1.85 & 53.87 $\pm$ 1.92 & 37.52 $\pm$ 1.52 & 79.33 $\pm$ 1.66 & 79.14 $\pm$ 0.95 \\
          % & \cmark & \cmark & \cmark & 91.x64 $\pm$ 2 & 95.x37 $\pm$ 3.x31 & \textbf{95.x25 $\pm$ 2.x37} & 40.x47 $\pm$ 1.x49 & 68.x93 $\pm$ 2.x04 & 54.x78 $\pm$ 1.x27 & \textbf{89.x13 $\pm$ 1.x77} & \textbf{81.x96 $\pm$ 2.x03} & \textbf{91.x01 $\pm$ 0.x7} & \multicolumn{1}{c}{3.x11} \\
    \hline
    \multicolumn{1}{|c|}{\multirow{3}[1]{*}{ACM w/}} & \xmark & \xmark &        94.43 $\pm$ 2.76 & 96.12 $\pm$ 2.05 & 95.57 $\pm$ 3.60 & 41.98 $\pm$ 0.89 & 68.45 $\pm$ 1.67 & 56.31 $\pm$ 1.23 & 88.56 $\pm$ 0.87 & 81.94 $\pm$ 1.48\\
          & \cmark &  \xmark     &  95.08 $\pm$ 2.43 & 97.38 $\pm$ 1.53 & 94.92 $\pm$ 2.25 & 43.05 $\pm$ 1.34 & 69.10 $\pm$ 1.87 & 56.20 $\pm$ 2.77 & 89.69 $\pm$ 1.08 & 81.39 $\pm$ 1.38 \\
          & \cmark & \cmark &   \textbf{95.41 $\pm$ 2.72} & \textbf{98.00 $\pm$ 1.50} & \textbf{96.72 $\pm$ 1.47} & \textbf{43.39 $\pm$ 1.39} & \textbf{69.87 $\pm$ 1.63} & \textbf{56.47 $\pm$ 1.36} & \textbf{89.89 $\pm$ 0.90} & \textbf{82.37 $\pm$ 1.09}  \\
    \hline
    % \bottomrule
    \end{tabular}%
    }
    % \vspace{-5mm}
\end{table*}%

We investigate the incremental model performance for all nodes on open-source datasets by segmenting test nodes based on the node homophily ratio $\mathcal{H}_{node}$, following \cite{udgnn}. High homophily indicates low noise and vice versa. Nodes with $\mathcal{H}_{node} \leq 0.25$ constitute the strong-heterophilous (SHet) segment, $0.25<\mathcal{H}_{node}\leq0.5$ the weak-heterophilous (WHet) segment, $0.5<\mathcal{H}_{node}\leq0.75$ the weak-homophilous (WHom) segment, and $0.75<\mathcal{H}_{node}\leq1.0$ the strong-homophilous (SHom) segment. Fig.~\ref{fig:sw_homo_hetero} demonstrates that $P^2$GNN improves performance on nodes with low levels of homophily (more noisy neighborhood), indicating that prototypes offer homophilous messages (positive signals) to heterophilous (noisy) nodes, resulting in better classification compared to ACM-GCN~\cite{luan2022revisiting}. \textit{Key takeaway: $P^2$GNN improves performance on noisy nodes by providing homophilous messages from prototypes, leading to better classification compared to ACM-GCN.}

\subsection{RQ3: Qualitative Analysis}
To comprehend the contribution and assess the desired objective of each prototype, we focus on the ACM-GCN+$P^2$GNN setup across small open-source datasets. We analyze: (i) the model's attention on prototypes, quantifying their relevance, (ii) the representation capability of prototypes through diversity and class-level concentration, and (iii) the change in embedding quality by incrementally adding prototype sets, visualized using t-SNE~\cite{JMLR:v9:vandermaaten08a}.

\textbf{Attention on Prototypes}: Fig.~\ref{fig:attention-plots} illustrates attention values ($\alpha_{P_\mathcal{N}}^L$) on test data, where x-axis represents attention values, and y-axis shows the percentage of nodes utilizing the corresponding attention value. In the top pane, all nodes have non-zero attention values, indicating effective use of neighboring prototypes to enhance node classification, evident from the accuracy increase for Chameleon (68.45\% to 69.10\%) and Cora (88.56\% to 89.69\%) datasets (Table~\ref{tab:ablation_study}). This validates usefulness of providing global information through prototypes. The bottom pane captures the mixing-attention values on $P_\mathcal{A}$ ($\alpha_{P_\mathcal{A}}^L$), exhibiting similar behavior, validating the use of prototypes for message alignment and performance improvement. Higher attention values are given to noisier dataset, showcasing the increased benefit of providing prototypes when the dataset is noisier. \textit{Key takeaway: Non-zero attention values on prototypes effectively demonstrate their utilization for performance improvement.}

\begin{figure*}[h]
  \centering
  \subfigure{\includegraphics[width=0.32\textwidth]{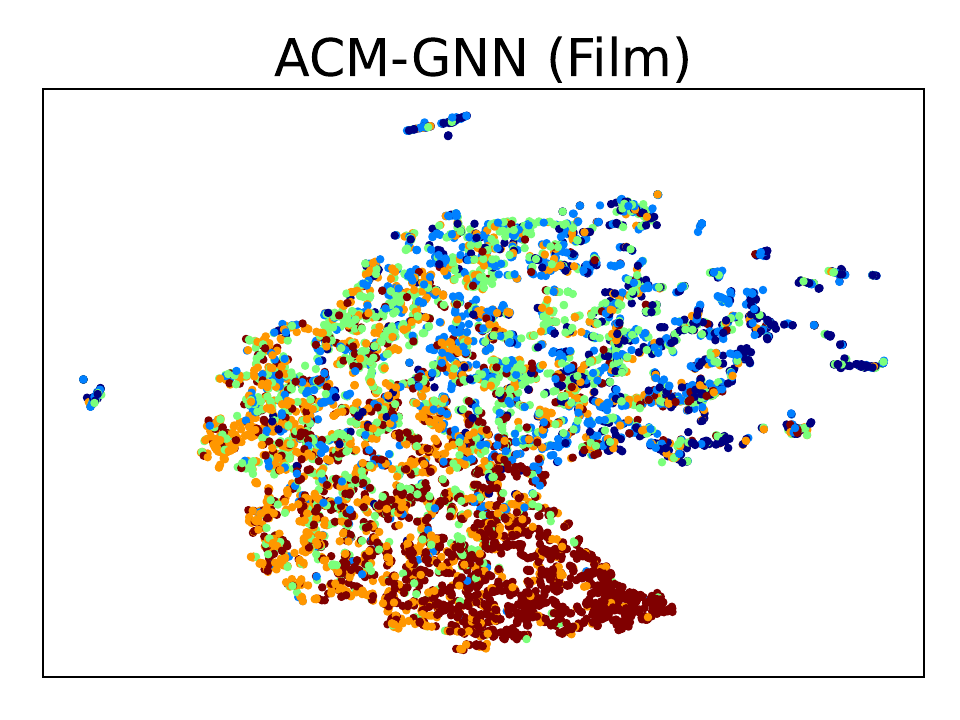}}
  \subfigure{\includegraphics[width=0.32\textwidth]{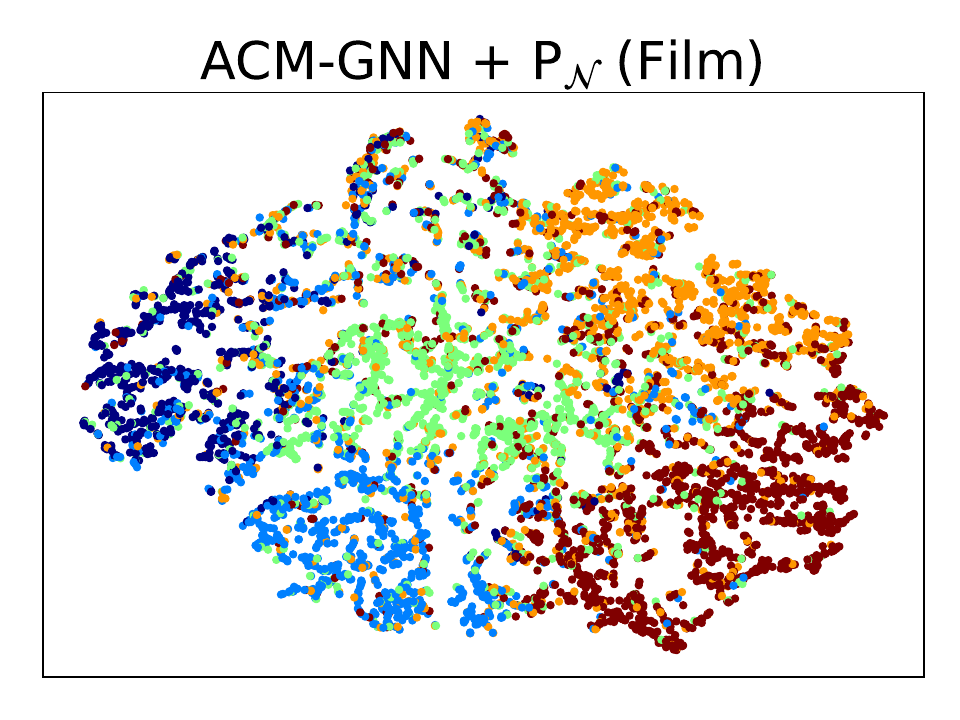}}
  \subfigure{\includegraphics[width=0.32\textwidth]{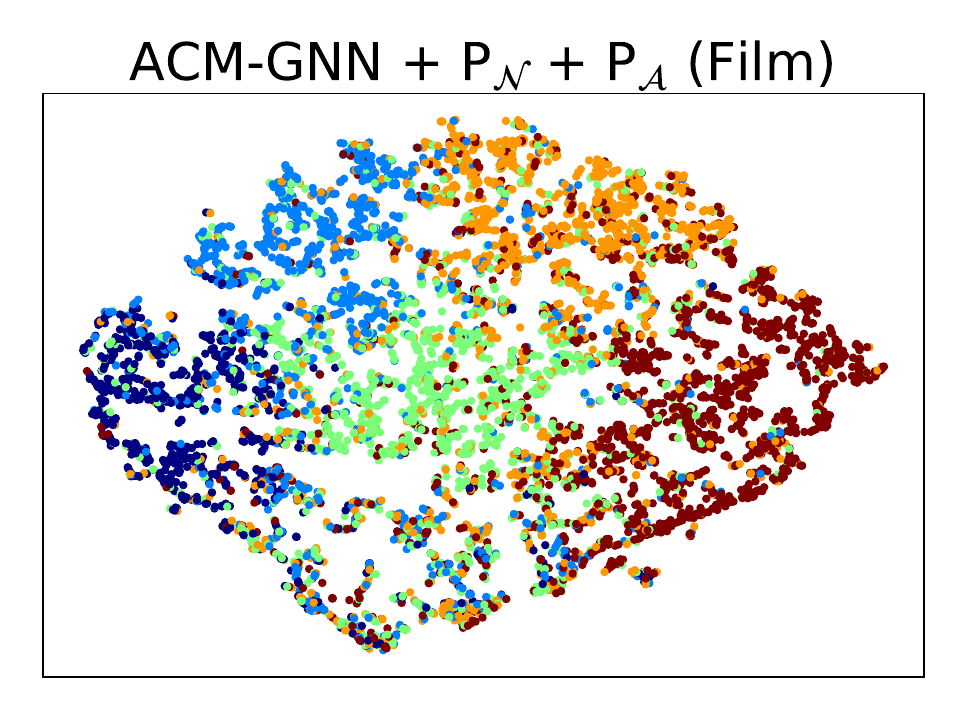}}
  \caption{tSNE plots of the penultimate layer embeddings of $P^2$GNN and the backbone ACM-GCN model. We progressively add the two different prototypes $P_{\mathcal{N}}$ and $P_{\mathcal{A}}$ and observed embeddings become well separated among the different classes (colors).}
  \label{fig:tsne-appendix}
\end{figure*}

\begin{figure*}[tp]
  \centering
  \subfigure{\includegraphics[width=0.19\textwidth]{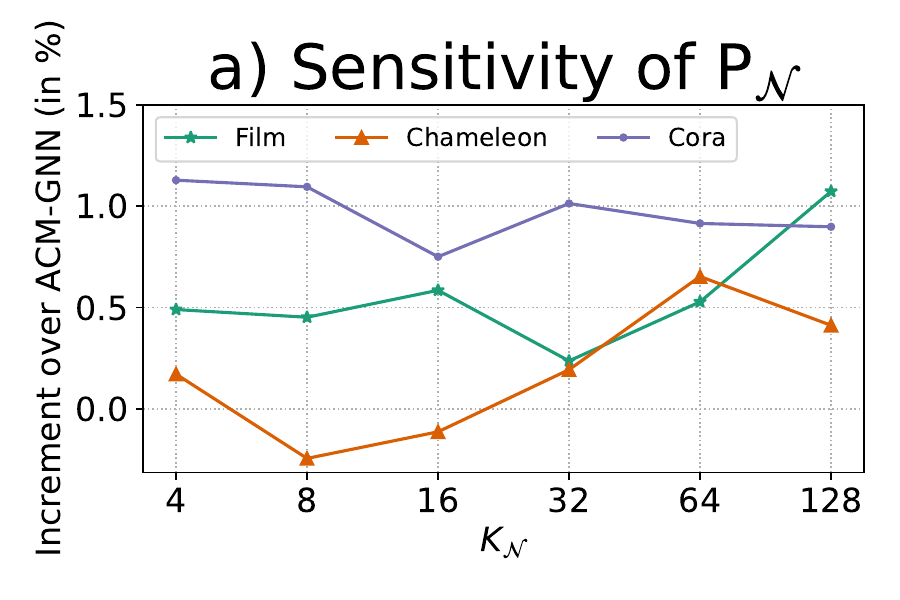}}
  \subfigure{\includegraphics[width=0.19\textwidth]{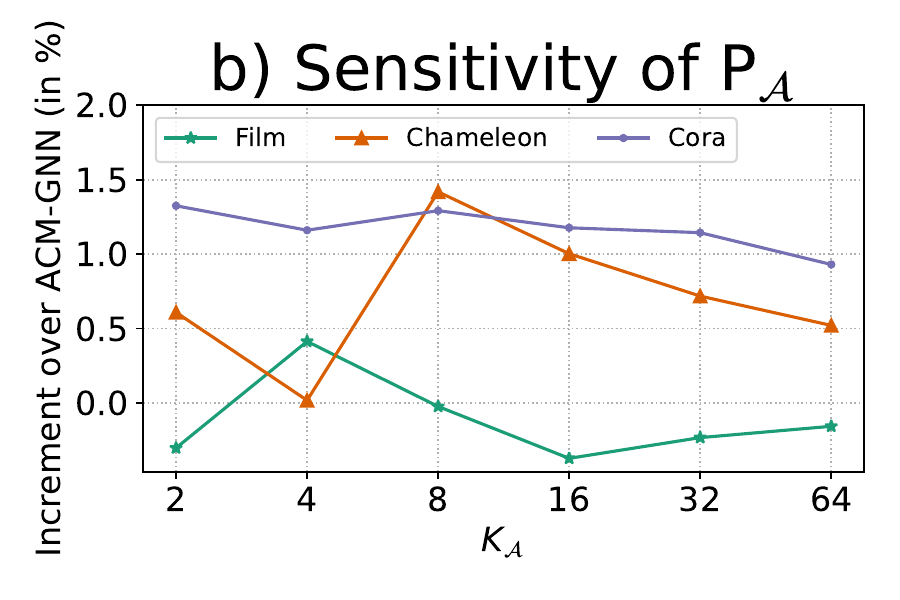}}
\subfigure{\includegraphics[width=0.19\textwidth]{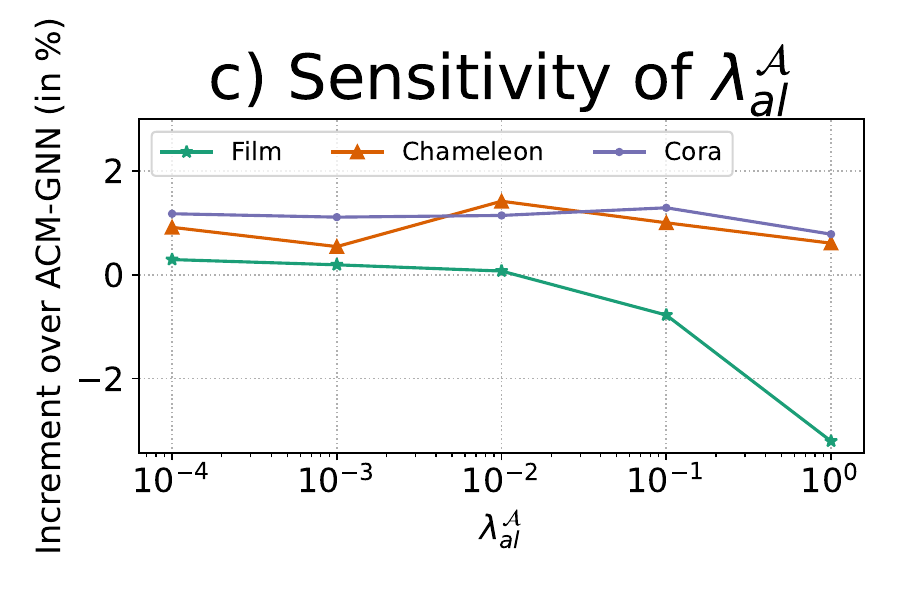}}
\subfigure{\includegraphics[width=0.19\textwidth]{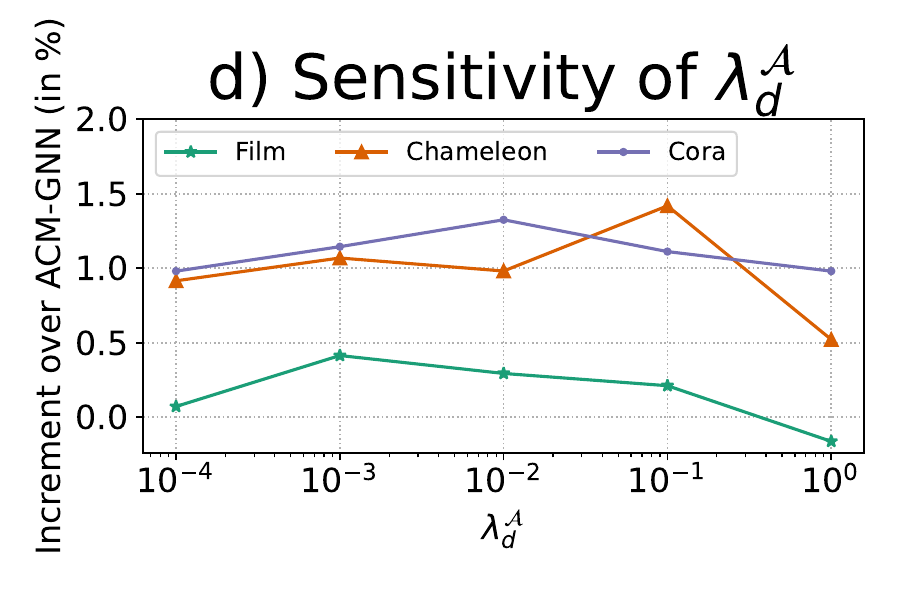}}
\subfigure{\includegraphics[width=0.19\textwidth]{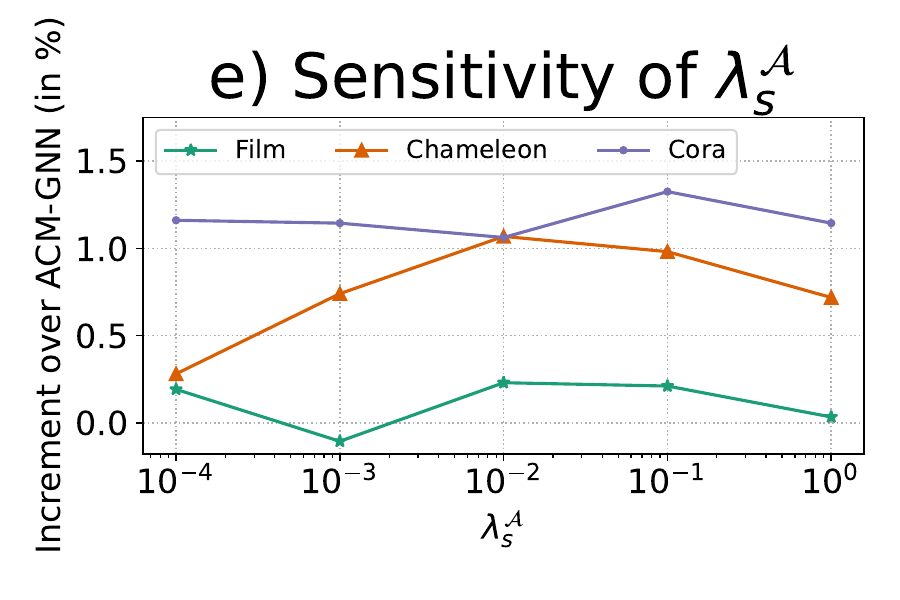}}
  \caption{Sensitivity of incremental performance of $P^2$GNN over ACM-GCN backbone towards different hyper-parameters. We experiment with different number of prototypes for $P_\mathcal{N}$ and $P_\mathcal{A}$ and relative weights of each of the 3 loss functions. While varying each hyper-parameter, the values of others are kept constant. }
  \label{fig:sensitivity}
\end{figure*}

\textbf{Prototype Representations:} We conduct a deeper analysis of the learned  $P_\mathcal{N}$ prototype representations to gain insights into their representation space. For this, we examine the first layer $P_\mathcal{N}$ representation, as they should reside in the same space as node features $X$. In Fig.~\ref{fig:p2gnn-representation}, we observe that prototypes are distinct, diverse, and span the entire input representation space, capturing global information. Notably, they do not collapse to a single representation value. \textit{Key takeaway: The learned prototypes effectively capture the global information and diversity trends present in the input data.}

\textbf{Embedding Quality}: Fig.~\ref{fig:tsne-appendix} presents t-SNE of penultimate layer embeddings for ACM-GCN, followed by one that includes only neighborhood prototypes (ACM-GCN+$P_\mathcal{N}$), and then with the addition of message-alignment prototypes (ACM-GCN+$P_\mathcal{N}$+$P_\mathcal{A}$). We observe progressive separability among embeddings from different classes, consistent with the denoising assumption from prototypes, as samples from the same class are better clustered in the embedding space when prototypes are present. \textit{Key takeaway: Adding prototypes improves the separability of learned embeddings providing denoising effect, leading to better class clustering.}

\subsection{RQ4: Hyperparameter Tuning and Sensitivity Analysis}
\label{sec:params}
The introduction of prototypes at two levels brings new hyperparameters, such as the number of prototypes at each level and the weights of the auxiliary loss functions. We varied the number of neighborhood prototypes ($K_\mathcal{N}$) and message-alignment prototypes ($K_\mathcal{A}$) among $\{2, 4, 8, 16, 32, 64\}$, and the loss weights ($\lambda_a, \lambda_d, \lambda_s$) among $\{1\mathrm{e}{-4}, 1\mathrm{e}{-3}, 1\mathrm{e}{-2}, 1\mathrm{e}{-1}, 1\}$. To find the optimal configuration efficiently, we performed Bayesian hyperparameter search, which is computationally more efficient than the extensive grid search employed by existing methods \cite{glognncode, acm-largecode, acm-smallcode} such as LINKX \cite{glognncode} and ACM-GNN \cite{acm-smallcode, acm-largecode} use nested loops for grid search.

\textbf{Hyperparameter Tuning Strategy:} In practice, we recommend starting the training process with high values for $K_\mathcal{N}$ and $K_\mathcal{A}$ and gradually reducing them to find the optimal set of hyperparameters. While very high values may provide some benefits, they may not necessarily yield the best performance for all datasets due to the risk of over-diversification or over-fitting. Therefore, starting with high values and decreasing them can help identify the most suitable choices efficiently. When tuning the loss weights $\lambda_a$, $\lambda_d$, and $\lambda_s$, it is important to ensure that their scale aligns with the scale of the cross-entropy loss used for the node classification task. Generally, using a value $\leq$ 0.1 resulted in stable training and improved performance. \textit{Key Takeaway: We recommend a hyperparameter tuning strategy suitable for different datasets to tune the model effectively.}

\textbf{Sensitivity to Number of Prototypes:} Fig.~\ref{fig:sensitivity} (a) highlights that the benefit of adding $P_{\mathcal{N}}$ prototypes depends on the average homophily (noise) ratio ($\mathcal{H}_{node}$) of the dataset. For low-homophily (high noise) datasets like Film and Chameleon, a larger number of neighborhood prototypes improves test accuracy by better capturing the diverse neighborhood patterns. Conversely, for high-homophily (low noise) datasets like Cora, the benefit stabilizes with a small number of prototypes. Fig.~\ref{fig:sensitivity} (b) shows that the right choice of $P_\mathcal{A}$ prototypes can boost performance, as message alignment heavily depends on the dataset's feature space. Too many message-alignment prototypes can lead to over-diversification of messages, similar to the elbow behavior in $K$-Means clustering. \textit{Key Takeaway: The optimal number of prototypes depends on the dataset's noise (homophily) characteristics and feature space.}

\textbf{Sensitivity to Auxiliary Loss Weights:} Fig.~\ref{fig:sensitivity} (c), (d), and (e) highlight the importance of tuning weights of auxiliary loss functions relative to the node classification cross-entropy loss. In each scenario, performance improvement is observed as relative weights increase, followed by a decline with further weight increase. This behavior is typical for auxiliary loss functions, as over-indexing on either alignment, diversity, or sparsity reduces the representational capability of prototypes, resulting in a drop in performance. \textit{Key Takeaway: Proper tuning of auxiliary loss weights is crucial to balance the different desiderata and achieve optimal performance.}

\section{Conclusion}

We addressed two critical challenges encountered by message passing GNNs: the absence of global information and the presence of noisy neighborhood representations. These factors often lead to suboptimal performance on certain datasets and render GNNs unreliable for industry applications. To mitigate these issues, we proposed a plug-and-play twin-prototype method seamlessly integrated with existing GNN architectures. Our approach leverages prototypes from two perspectives and employs carefully crafted loss functions to enhance model performance and embedding quality. Demonstrating superior performance over SOTA baselines across datasets and tasks, our method solidifies its position as a leading technique. Extensive qualitative analysis further validated our intuition and hypotheses regarding the efficacy of prototype integration, offering insights into the drivers behind enhanced model performance. Notably, the proposed methodology is generic and readily adaptable to most GNN layers, as outlined in Fig.~\ref{fig:schematic} and Algorithm~\ref{alg:additionP2}. Furthermore, it holds promise for extension into emerging directions within the realm of Graph Machine Learning.

\bibliographystyle{plainnat}
\bibliography{references}

\appendix

\section{Dataset Details}
\label{sec:datasets}
\begin{table}[htbp]
    \centering
    \caption{Statistics of different datasets used for experimentation. \# C is the number of distinct node classes. $\mathcal{H}_{node}$ measures the average proportion of label consistency between a node and its neighbors.}
    % \vspace{-3mm}
    \label{tab:data_stats}
    {\footnotesize
    \begin{tabular}{ccrrrrrrr}
    \toprule
    
    Type & Dataset &  \# Nodes & \# Edges &  \# Feat. & \# C & $\mathcal{H}_{node}$ \\
    \midrule
    \multirow{8}{*}{Small} & Texas  & 183 & 309 & 1,703 & 5 & .11 \\
         & Wisconsin & 251 & 499 & 1,703 & 5 & .21 \\
         & Cornell & 183 & 295 & 1,703 & 5 & .30 \\
         & Film & 7,600 & 29,926 & 931 & 5  & .22 \\
         & Chameleon & 2,277 & 36,101 & 2,325 & 5 & .23 \\
         & Squirrel & 5,201 & 216,933 & 2,089 & 5 & .22 \\
         & Cora & 2,708 & 5,278 &  1433 & 7 &  .81 \\
         & Citeseer & 3,327 & 4,552 & 3703 & 6 &  .74 \\
         
    \midrule
         \multirow{6}{*}{Large} & Penn94 & 41,554 & 1,362,229  & 5 & 2  & .470  \\
         & pokec  & 1,632,803 & 30,622,564 & 65  &  2  & .445 \\
         & arXiv-year  & 169,343 & 1,166,243 & 128 & 5& .222 \\
         & snap-patents  & 2,923,922 & 13,975,788 & 269 & 5& .073  \\
         & genius  & 421,961 & 984,979 & 12 & 2& .618  \\
         & twitch-gamers  & 168,114 & 6,797,557 & 7 & 2& .545 \\
         
    \midrule
    
        \multirow{2}{*}{Fraud} & YelpChi & 45,954 & 3,846,979 & 32 & 2 & .227 \\
        & Amazon & 11,944 & 4,398,392 & 25 & 2 & .046 \\
        \midrule
        \multirow{2}{*}{Proprietary} & E-comm1 & 3.5M & 404M & 384 & - & - \\
        & E-comm2 & 686K & 10.5M & 768 & - & - \\
    \bottomrule
    \end{tabular}
    }
    % \vspace{0pt}
\end{table}
Table~\ref{tab:data_stats} mentions the details of different datasets used for experimentation.
\begin{table*}[t]
\centering
%\resizebox{0.87\linewidth}{!}{
\caption{The classification accuracy $\pm$ standard deviation (in \%) over the methods on 8 small-scale datasets is reported on the fixed 48\%/32\%/20\% data split. We highlight the best score on each dataset in \textbf{bold} and the runner-up score with \underline{underline}.
% \vspace{-3mm}
% Note that Edge Hom.~\cite{zhu2020beyond} is defined as the fraction of edges that connect nodes with the same label.
}
\label{tab:ours_vs_sota_fixed}
\resizebox{\linewidth}{!}
{
\begin{tabular}{|c|c|c|c|c|c|c|c|c||c|}
\hline
                    & \textbf{Texas} & \textbf{Wisconsin}     & \textbf{Cornell} & \textbf{Film} & \textbf{Squirrel} & \textbf{Chameleon} & \textbf{Cora} & \textbf{Citeseer}  & \textbf{Avg. Rank}
                    % \parbox[t]{2mm}{\multirow{5}{*}{\rotatebox[origin=c]{90}{\textbf{Avg. Rank}}}} 
                    \\
$\mathcal{H}_{node}$  &   0.11    &      0.21     &  0.30  &  0.22  & 0.22  &  0.23 &  0.81 &  0.74 &   \\
% \textbf{\#Nodes}  &   183    &   251        &  183  &  7,600  & 5,201 & 2,277  & 2,708 & 3,327 &  19,717 &  \\
%  \textbf{\#Edges}  &    295   &    466       &  280  &  26,752  & 198,493 & 31,421  &  5,278 & 4,676  &  44,327 &   \\
%  \textbf{\#Features}  &   1,703    &    1,703       &  1,703   &  931  &  2,089 &  2,325 &  1,433 & 3,703  & 500  &   \\
%  \textbf{\#Classes} &    5   &     5      & 5  &  5  & 5 & 5  & 6 & 7  &  3  & \\ 
\hline   
     MLP                   &  $80.81 \pm 4.75$  &     $85.29\pm 3.31$        &  $81.89 \pm 6.40 $  &  $36.53 \pm 0.70$  &  $28.77 \pm 1.56$ & $46.21 \pm 2.99 $  & $75.69 \pm 2.00 $  & $74.02 \pm 1.90$  &  12.4 \\
     GCN~\cite{kipf2016classification}                    &  $55.14 \pm 5.16$    &  $51.76 \pm 3.06$          & $60.54 \pm 5.30$ & $27.32\pm 1.10$ &   $53.43\pm 2.01$  & $64.82 \pm 2.24$ & $86.98 \pm 1.27$ &  $76.50 \pm 1.36$    &  13.0 \\
     GAT~\cite{velivckovic2017attention}                    &  $52.16 \pm 6.63$   &   $49.41 \pm 4.09$          &$61.89 \pm 5.05$ & $27.44 \pm 0.89$ &  $40.72 \pm 1.55$ &  $60.26 \pm 2.50$  & $ 87.30 \pm 1.10$ & $76.55 \pm 1.23$ &   13.4 \\    
    %   MixHop~\cite{abu2019mixhop}               &   $77.84 \pm 7.73$   &     $ 75.88\pm 4.90$       & $73.51 \pm 6.34$ & $ 32.22\pm 2.34$ & $43.80\pm 1.48$ & $ 60.50\pm 2.53$  & $ 87.61\pm 0.85$ & $ 76.26\pm 1.33$ &  12.375   \\  
%     GraphSAGE     & $82.43 \pm 6.14$  &  $81.18 \pm 5.56$           & $75.95\pm 5.01$& $34.23\pm 0.99$ &$41.61\pm 0.74 $  &  $58.73\pm 1.68$  & $86.90\pm 1.04$ &  $76.04 \pm 1.30$ & $88.45 \pm 0.50$ &    \\
%     PairNorm                    &   $60.27\pm 4.34$   &   $48.43 \pm 6.14$         & $58.92 \pm 3.15$ & $27.40 \pm 1.24$ & $50.44\pm 2.04$ & $62.74 \pm 2.82$  & $85.79 \pm 1.01$ & $73.59 \pm 1.47$ &  $87.53 \pm 0.44$  &   \\
 %    Geom-GCN                  &  $66.76 \pm 2.72$   &    $64.51 \pm 3.66$         &$60.54\pm 3.67$ & $31.59\pm 1.15$ & $38.15\pm 0.92$ &  $60.00 \pm 2.81$  & $85.35 \pm 1.57$ & $\bm{78.02 \pm 1.15}$ &  $\underline{89.95 \pm 0.47}$  & \\   
   %  FAGCN                &  $ \pm $   &    $ \pm $         &$\pm $ & $\pm $ & $\pm $ &  $ \pm $  & $ \pm $ & $ \pm $ &  $\pm $  & \\     
     
    H$_2$GCN~\cite{zhu2020beyond}               &  ${84.86 \pm 7.23}$    &    ${87.65 \pm 4.98}$        & $82.70 \pm 5.28$ & $35.70 \pm 1.00$ & $36.48 \pm 1.86$ & $60.11\pm 2.15$  & $87.87\pm 1.20$ & $77.11\pm 1.57$    &  9.0 \\ 
    WRGAT~\cite{suresh2021breaking}               &  $ 83.62 \pm 5.50 $    &    ${ 86.98 \pm 3.78}$        & $ 81.62 \pm 3.90$ & $ 36.53 \pm 0.77 $ & $ 48.85 \pm 0.78 $ & $65.24 \pm 0.87$  & $88.20 \pm 2.26$ & $76.81 \pm 1.89 $ & 8.3  \\ 
    GPR-GNN~\cite{chien2021adaptive}              &   $78.38\pm 4.36$   &     $82.94 \pm 4.21$       & $80.27 \pm 8.11$ & $34.63 \pm 1.22$ & $31.61 \pm 1.24$ & $46.58\pm 1.71$  & $87.95 \pm 1.18$ & $77.13 \pm 1.67$ &  11.3  \\

       %ACM-GCN               &  $   \pm $   &     $   \pm  $       & $  \pm  $ & $  \pm  $ & $   \pm  $ & $  \pm  $  & $  \pm  $ & $   \pm  $ & $  \pm $  &    \\  
       LINKX~\cite{lim2021large}               &   $ 74.60 \pm 8.37$   &     $75.49  \pm 5.72$       & $ 77.84 \pm 5.81$ & $ 36.10\pm 1.55$ & ${61.81  \pm 1.80}$ & $ 68.42 \pm 1.38$  & $ 84.64 \pm 1.13$ & $ 73.19 \pm 0.99 $ &  11.4  \\ 
    GloGNN~\cite{li2022finding}               &   $ 84.32 \pm 4.15$   &     $87.06 \pm 3.53$       & $ 83.51 \pm 4.26$ & $ 37.35\pm 1.30$ & $ 57.54 \pm 1.39$ & $ {69.78 \pm 2.42}$  & $ \underline{88.31 \pm 1.13}$ & $ \underline{77.41 \pm 1.65}$ &  5.3  \\

    % Ordered GNN & 86.22 $\pm$ 4.12 & 88.04 $\pm$ 3.63 & 87.03 $\pm$ 4.73 & 37.99 $\pm$ 1.00 & 62.44 $\pm$ 1.96 & 72.28 $\pm$ 2.29 & 88.37 $\pm$ 0.75 & 77.31 $\pm$ 1.73 & & \\
    
    ProtoGNN~\cite{dongprotognn} & $81.08\pm6.24$ & $86.07\pm5.22$ & $79.27\pm4.54$ & $30.62\pm0.07$ & $59.29 \pm 1.70$ & $64.23 \pm 2.69$ & $82.55\pm1.39$ & $72.71\pm1.35$ & 11.5\\
    
    Dual-Net GNN~\cite{maurya2022not} & $84.59\pm3.83$ & $86.96\pm3.79$ & \underline{$86.08\pm5.75$} & $36.91\pm0.74$ & \bm{$73.65 \pm 1.97$} & $78.20 \pm 1.07$ & $86.91\pm1.42$ & $76.68\pm1.69$ & 5.9\\

      HP-GMN~\cite{hpgmn} & $85.10 \pm 4.20$ & $86.51 \pm 3.10$ & $84.10 \pm 5.30$ & NA & $\underline{72.30 \pm 2.30}$ & $\bm{79.60 \pm 0.80}$ & $84.70 \pm 0.60$ & $73.00 \pm 1.00$ & 8.1\\
    
      GloGNN++~\cite{li2022finding}               &   $ 84.05  \pm 4.90$   &     \underline{$88.04 \pm 3.22$}       & $ {85.95 \pm 5.10}$ & $ \bm{37.70 \pm 1.40}$ & ${57.88 \pm 1.76}$ & $ {71.21 \pm 1.84}$  & $ \bm{88.33 \pm 1.09}$ & $ 77.22 \pm 1.78$ &  $\underline{4.0}$   \\
    
    GCN\rom{2}~\cite{chen2020simple}                 & $77.57 \pm 3.83$  &   $80.39 \pm 3.40$          & $77.86 \pm 3.79$ &  $37.44 \pm 1.30$  & $38.47 \pm 1.58$ & $63.86 \pm 3.04$  & $88.37 \pm 1.25$ & $77.33 \pm 1.48$ &  9.0 \\
    
    GGCN~\cite{yan2021two}              &   $84.86 \pm 4.55$   &     $86.86\pm 3.29$       & $85.68 \pm 6.63$ & $ \underline{37.54 \pm 1.56}$ & $55.17\pm 1.58$ & $71.14 \pm 1.84$  & $87.95 \pm 1.05$ & $77.14 \pm 1.45$ &  5.4 \\ 
    
    ACM-GCN~\cite{luan2022revisiting}               &  $ \underline{87.84  \pm 4.40}$   &     \bm{$  88.43 \pm 3.22$}       & $ 85.14 \pm 6.07 $ & $ 36.28 \pm 1.09 $ & $ 54.40  \pm 1.88 $ & $ 66.93 \pm 1.85 $  & $ 87.91 \pm 0.95 $ & $ 77.32  \pm 1.70 $ &   5.8   \\

      \hline
      $P^2$GNN & $\bm{88.65 \pm 4.80}$ & \bm{$ 88.43 \pm 3.09$} & $\bm{86.49 \pm 6.04}$ & $ 37.47 \pm 1.11$ & $71.88 \pm 1.76$ & \underline{${78.60 \pm 1.41}$} & $88.23 \pm 1.07$ & $\bm{77.51 \pm 1.45}$ & $\bm{2.0}$ \\
      \hline

\end{tabular}
}
\end{table*}

\section{Implementation Details}
\label{sec:implementaion}
We implemented the experiments using DGL~\cite{wang2019dgl} and PyTorch. For small and large datasets, we followed the same setup as ACM-GCN, utilizing the same learning rate, optimizer, and number of GNN layers. A similar setup was used for the fraud detection dataset, following the GHRN~\cite{gao2023addressing} setup. For the proprietary dataset, we employed a 2-layer GCN as the backbone and used the parameters mentioned in Section~\ref{sec:params} for tuning prototypes. Experiments on small and fraud datasets were conducted on a 64-core machine with 488 GB CPU RAM running Linux, while experiments on large and proprietary datasets were performed on 8 V100 GPUs with 32 GB RAM each.

\section{Additional Results}
\label{sec:48split}
Table~\ref{tab:ours_vs_sota_fixed} shows the test classification accuracy of $P^2$GNN compared with various SOTA on fixed splits datasets as specified in \cite{pei2020geom}. Our approach achieved the highest rank among all other approaches, demonstrating its efficacy independent of the training data split. $P^2$GNN also outperforms existing prototype-based methods, namely ProtoGNN~\cite{dongprotognn} and HP-GMN~\cite{hpgmn}. For completeness, we attempted to experiment with HP-GMN on the Film datasets using the code\footnote{https://github.com/junjie-xu/HP-GMN}, but were unable to achieve comparable performance. This issue has already been reported on their repository\footnote{https://github.com/junjie-xu/HP-GMN/issues/1} but has not been addressed yet, raising concerns regarding the reproducibility of their results.

\section{Design Choices for Prototypes}
Intuitively, the prototypes should summarise different types of nodes present in the graph, where the "types" are not restricted to the different classes present in graph, to accommodate for high intra-class variance. Our prototypes ($P_N$) aim at only increasing homophilous message exposure to nodes in the feature propagation step, which is possible by connecting the prototypes to all the nodes in the graph. This incremental connection aims to leverage the strength of aggregation methodologies that focus on homophilous messages (eg: GATv2~\cite{velivckovic2017attention}, ACM-GCN~\cite{luan2022revisiting}) and reduce the impact of messages from heterophilous edges (i.e. heterophilous prototypes). We can also leverage techniques like GBK-GNN~\cite{du2022gbk}, CARE-GNN~\cite{dou2020enhancing} as backbones to minimize the impact of heterophilous edges coming from prototypes. 
~\cite{dongprotognn} is a recent work that works with class-dependent prototypes, restricting its scope to binary classification tasks and performs poorer with respect to $P^2$GNN (ref. Table~\ref{tab:ours_vs_sota_fixed}). 

However, the idea behind prototypes ($P_N$) is to increase exposure of homophilous messages to all nodes, hence we added connected them to all nodes. The addition of edges from prototypes to nodes based on known class-labels of prototypes will lead to label-leakage. ProtoGNN~\cite{dongprotognn} also uses all the prototypes (independent of their class) for the slot-attention mechanism. 

It is important to note that during message-aggregation we treat the messages from neighbours and prototypes ($P_N$) differently by obtaining separate attention values of $\alpha^l_{base}$, $\alpha^l_{P_N}$. Addition of edges from prototypes to all nodes and aggregating messages from neighbouring nodes and prototypes is similar in spirit to the mechanism of performing GNN aggregation from connect neighbours followed by using prototypes in slot-attention mechanism of ProtoGNN. Thus any method that would use prototypes will either connect them to all nodes (in some form or the other) or perform selective edge-addition.

\end{document}